# An Auto-scaling approach for serverless environments based on a multi-expert consensus mechanism

Mobina Kashaniyan[1], Mehrdad Ashtiani[*2], Amirhossein Ghassemi[3]

[1] School of Computer Engineering, Iran University of Science and Technology, Tehran, Iran
mobina_kashanian@comp.iust.ac.ir

[2] School of Computer Engineering, Iran University of Science and Technology, Tehran, Iran
m_ashtiani@iust.ac.ir

[3] School of Computer Engineering, Iran University of Science and Technology, Tehran, Iran
a_ghassemi@comp.iust.ac.ir

## Abstract

Serverless computing offers automatic resource management and pay-per-use execution, but autoscaling remains difficult due to cold-start latency, inter-function dependencies, and highly dynamic workloads. Many existing approaches scale functions independently or rely on a single predictor, which can reduce robustness and cost efficiency. We present a dependency-aware autoscaling framework that unifies bottleneck identification, short-horizon demand forecasting, and cost-aware control in an end-to-end pipeline. We model applications as directed dependency graphs and prioritize high-impact functions using degree centrality. For these bottlenecks, near-term demand is predicted using lightweight supervised models, whose outputs are fused via a performance-weighted probabilistic ensemble inspired by Bayesian model averaging to improve stability under workload variability. The controller also accounts for cold starts and filters candidate actions through a cost-comparison mechanism to balance latency and operational efficiency. Experiments on real workload traces show improved prediction accuracy and more stable scaling decisions than representative baselines; supervised forecasting also consistently outperforms unsupervised clustering for generating autoscaling actions. The primary contribution is a practical system-level design that integrates dependency analysis, ensemble-based prediction, and cost-aware decision-making for robust serverless autoscaling.

## 1. Introduction

Serverless computing allows developers to deploy event-driven functions without managing infrastructure, offering automatic provisioning and pay-per-use billing. Its fine-grained autoscaling adjusts function instances to workload variations, but accurate control remains challenging due to bursty demand, cold-start latency, and inter-function dependencies [1–4]. Unlike VM- or container-based systems, serverless applications often execute as chains of dependent function invocations. Congestion at one function can propagate along execution paths, increasing end-to-end latency and cost [5]. Autoscalers that treat functions independently may overlook structurally critical bottlenecks, while aggressive scale-down can increase cold starts and degrade performance. To address these challenges, we propose a dependency-aware autoscaling framework that integrates structural bottleneck identification, multi-model demand forecasting, cold-start awareness, and cost-comparison control within a unified pipeline. Workflows are represented as directed dependency graphs, and bottleneck functions are prioritized using degree centrality. Near-term demand for these functions is predicted using lightweight supervised models (MLP, LSTM, and CNN), and predictions are combined via a performance-weighted probabilistic ensemble inspired by Bayesian model averaging to reduce model-

[*] Corresponding author. Address: School of Computer Engineering, Iran University of Science and Technology, Hengam St., Resalat Sq., Tehran, Iran, Postal Code: 16846-13114, Fax: +98-21-73021480.

**Published in the Journal of Ambient Intelligence and Smart Environments**

Kashaniyan M, Ashtiani M, Ghassemi A. An auto-scaling approach for serverless environments based on a multi-expert consensus mechanism. Journal of Ambient Intelligence and Smart Environments. 2026 Jun 23:18761364261459585.

specific bias and improve stability. The controller also incorporates cold-start awareness and a cost-comparison step to balance latency and operational efficiency. We evaluate the framework on real workload traces and compare it with standard autoscaling baselines and learning-based approaches. Results show improved prediction accuracy and more stable scaling behavior, and a unified comparison indicates that supervised forecasting substantially outperforms unsupervised clustering for autoscaling decision generation.

The main contributions of this work are as follows:

1. **Dependency-aware autoscaling pipeline:** An end-to-end controller that couples dependency analysis, forecasting, and cost-aware actuation rather than scaling each function independently.
2. **Graph-based bottleneck identification:** A directed dependency graph with degree-centrality ranking to construct a watch set of high-impact functions for targeted monitoring and control.
3. **Lightweight multi-model forecasting:** Demand for bottleneck functions is predicted using three lightweight neural network models: multilayer perceptron (MLP), long short-term memory (LSTM), and convolutional neural network (CNN). These models capture complementary workload patterns and remain computationally suitable for online control.
4. **Bayesian-inspired consensus decision-making**: A probabilistic ensemble combines predictions from multiple models using validation-based weighting to improve stability and reduce model-specific bias.
5. **Cold-start aware and cost-comparison scaling:** The scaling policy takes into account cold-start considerations and aims to control costs effectively.

The remainder of the paper is organized as follows. Section 2 provides essential background. Section 3 reviews related work and identifies open gaps. Section 4 describes the proposed framework and implementation details. Section 5 reports experimental results and evaluation. Section 6 concludes the paper, and Section 7 presents future research directions.

## 2. Background Knowledge

This section briefly introduces the concepts necessary to understand the proposed autoscaling framework. We summarize the serverless execution model, describe dependency modeling through graph representation, and outline the learning paradigms used for demand prediction.

### 2.1 Serverless Execution Model

Serverless computing is commonly implemented through Function-as-a-Service (FaaS), where the cloud provider manages provisioning, scaling, and billing of function instances [6,7]. As shown in Figure 1, responsibility progressively shifts from the customer to the provider across deployment models, from on-premises to infrastructure-as-a-service (IaaS), platform-as-a-service (PaaS), function-as-a-service (FaaS), and software-as-a-service (SaaS). In FaaS, developers primarily manage application logic and data, while the provider controls the runtime, operating system, virtualization layer, and hardware. Serverless functions are typically stateless and event-driven, and are billed according to execution time and allocated memory [8,9]. A distinguishing feature is scale-to-zero capability, which releases resources during idle periods. While this improves cost efficiency, reactivating a function may introduce cold-start latency due to runtime initialization [10]. These characteristics make autoscaling both fine-grained and highly dynamic, requiring decisions that balance responsiveness, latency, and cost under rapidly changing workloads.

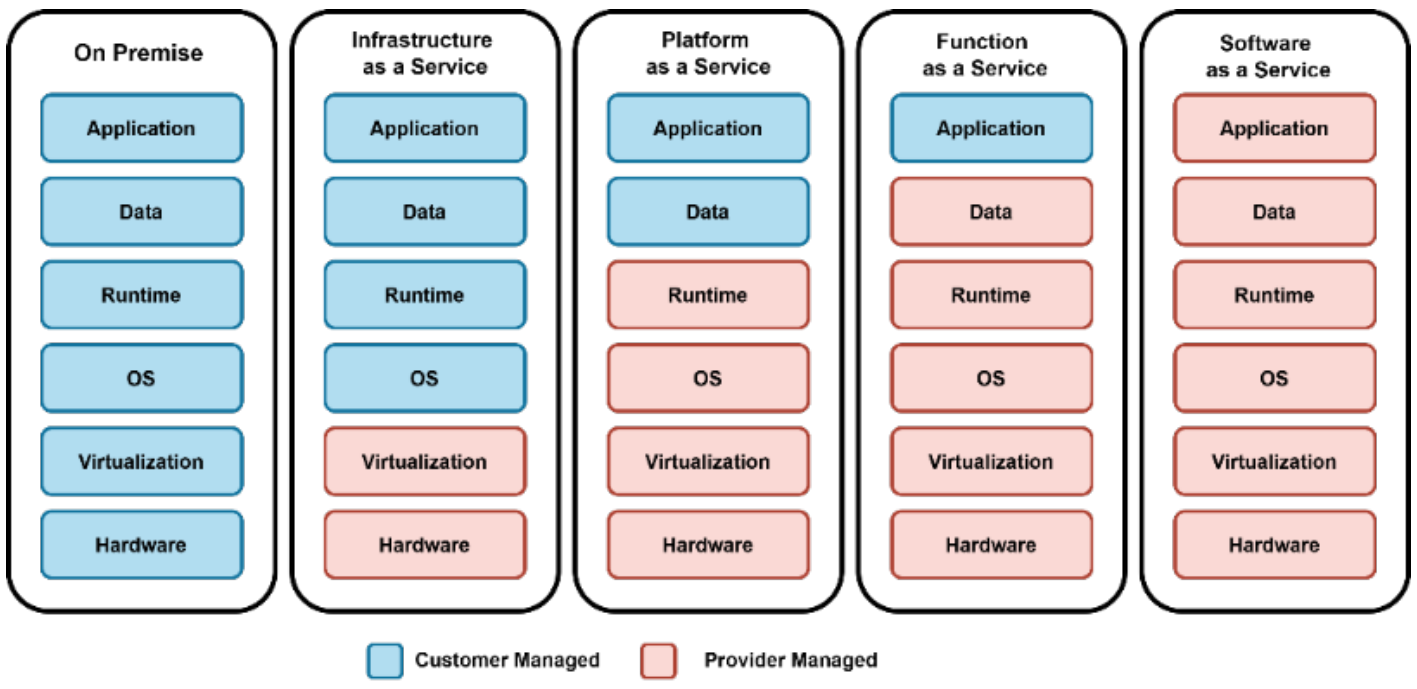

**Figure 1.** Responsibility distribution across cloud service models. Blue indicates customer-managed components and red indicates provider-managed components. In FaaS, users manage primarily application logic, while the provider controls the underlying infrastructure [11].

## 2.2. Dependency Modeling

Serverless applications commonly consist of multiple interacting functions connected through invocation relationships. A single request may traverse several functions, forming execution paths whose combined behavior determines end-to-end latency and resource consumption. To capture this structure, we represent the application as a directed dependency graph, where nodes correspond to functions and edges represent invocation dependencies [12]. This abstraction enables analysis of structural influence across the workflow. Graph analysis can identify functions that occupy critical positions within execution paths and are therefore more likely to accumulate workload pressure or propagate latency [13]. Prioritizing these structurally influential functions provides a principled basis for targeted autoscaling decisions.

## 2.3. Learning-Based Demand Prediction

Learning-based approaches are widely used in autoscaling to estimate near-term resource demand under dynamic workloads. In this work, supervised models generate actionable scaling decisions, while unsupervised techniques are evaluated only as comparative baselines.

### 2.3.1. Unsupervised learning (Baselines)

Unsupervised methods identify patterns in data without labeled outputs. Principal Component Analysis (PCA) reduces dimensionality by projecting data onto directions that capture maximum variance [14–16]. Clustering techniques, such as k-means, partition observations into similarity-based groups [17–20], while Self-Organizing Maps (SOM) provide topology-preserving clustering and visualization [21–23]. We also use t-SNE for the visualization of high-dimensional workload patterns [24,25]. Although these techniques reveal structural patterns in resource traces, they do not directly produce stable real-time scaling decisions. Therefore, they are used only to evaluate whether pattern discovery alone can approximate autoscaling actions.

### 2.3.2. Supervised Learning (Decision Models)

Supervised learning models learn mappings from observed resource metrics to future demand or scaling actions. Multilayer Perceptrons (MLP) capture nonlinear relationships through fully connected layers [26–28]. Long Short-Term Memory networks (LSTM) model temporal dependencies in sequential workload data [29–35]. Convolutional Neural Networks (CNN) extract local patterns from sliding windows of time-series inputs [36–39]. In the proposed framework, these complementary models serve as candidate predictors, and their outputs are later combined using Bayesian-inspired model averaging to improve robustness.

## 2.4. Auto-scaling Overview

Autoscaling dynamically adjusts compute resources to maintain performance objectives while controlling operational cost. The control granularity varies across platforms, including virtual machines and containers in instance-based clouds [40–43], service replicas in microservices architectures [44,45], and individual function instances in serverless systems [46]. Figure 2 presents a taxonomy view of the proposed framework across four dimensions: objectives, information sources, core modeling choices, and control characteristics. The framework aims to achieve accurate demand prediction, robust consensus-based decision-making, cold-start-aware behavior, and cost comparison-based scaling. Scaling decisions are guided by runtime signals and workflow dependencies; demand is forecast using supervised models aggregated via Bayesian-inspired consensus; and candidate scaling actions are evaluated through a cost-comparison component before execution.

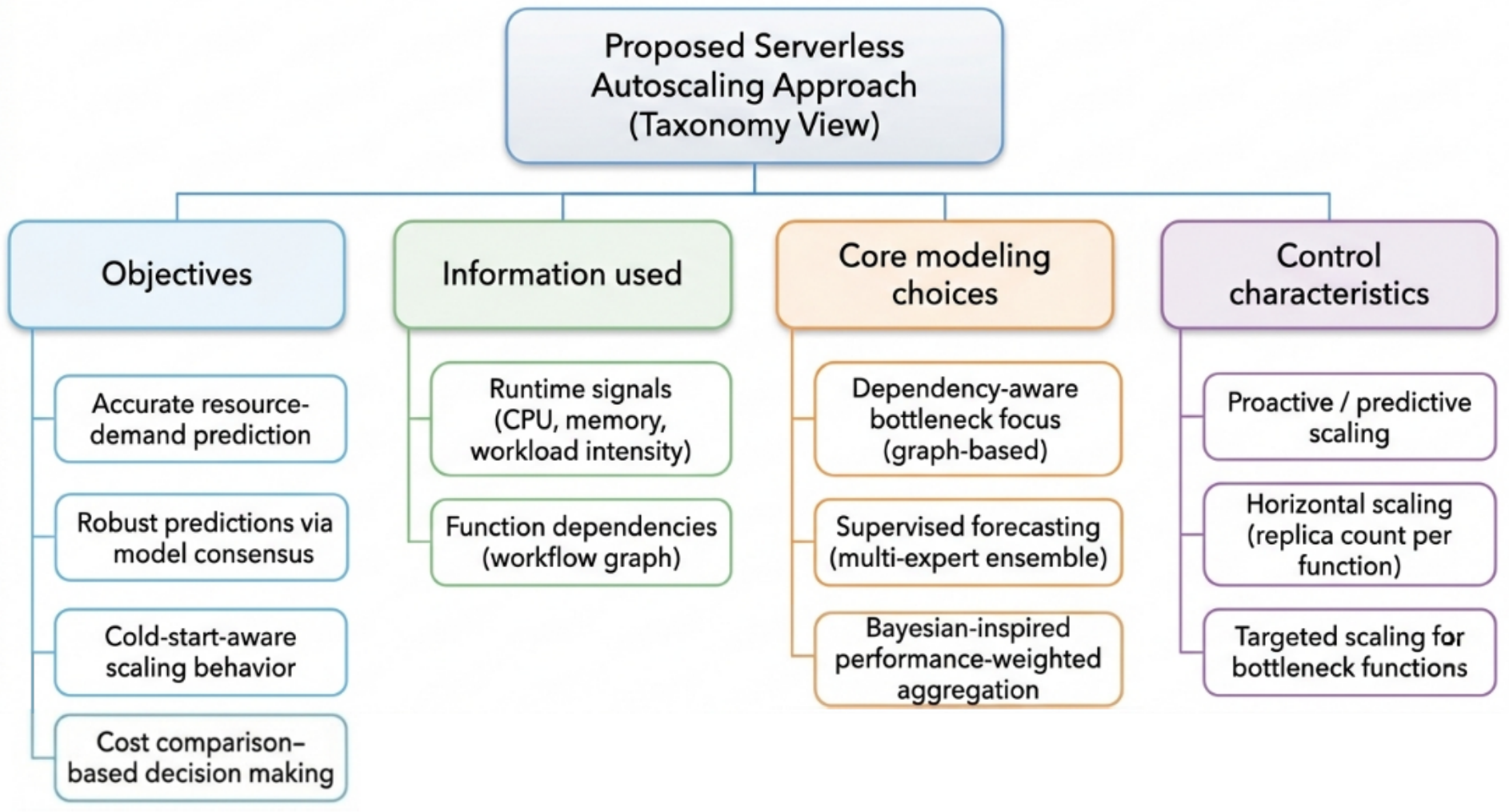

**Figure 2**. Taxonomy view of the proposed serverless autoscaling approach.

Autoscaling approaches are commonly categorized by how decisions are generated. Threshold-based approaches trigger scaling when monitored metrics exceed predefined limits [47]. Control-theoretic methods regulate system outputs toward target values using feedback mechanisms [48]. Learning-based approaches use supervised prediction or reinforcement learning to select scaling actions [49–52]. Queueing-based models estimate delay and required capacity from traffic characteristics [53,54]. Time-series forecasting methods predict short-horizon demand from historical observations [55,56]. Although these categories differ in modeling assumptions and responsiveness, serverless environments introduce additional constraints, including cold-start latency and inter-function dependencies. These characteristics motivate autoscaling designs that integrate predictive modeling with structural awareness, as reflected in the proposed framework.

## 2.5. Bayesian-Inspired Model Averaging

Ensemble methods combine predictions from multiple models to improve robustness and generalization performance [57]. Unlike methods that select a single best model, Bayesian Model Averaging (BMA) accounts for model uncertainty by forming a probabilistic weighted ensemble [58,59]. BMA has been successfully applied in regression, classification, and time-series forecasting tasks. In this study, BMA is used to produce consensus resource-demand predictions. Let $M_i$ denote candidate models and $D$ denote the observed data. The posterior probability of the model $M_i$ given data $D$ is computed using Bayes' theorem as defined in Equation 17:

$$P(M_i|D) = \frac{P(D|M_i)\, P(M_i)}{P(D)} \tag{17}$$

Where $P(D|M_i)$ is the likelihood of the data under the model $M_i$ and $P(M_i)$ is the prior probability of the model [60,61]. The marginal likelihood is described in Equation 18:

$$P(D) = \sum_i P(D|M_i)\, P(M_i) \tag{18}$$

The BMA prediction is obtained as the posterior-weighted average of individual model predictions, as shown in Equation 19:

$$y_{BMA} = \sum_{i} P(M_i|D)\, y_i \tag{19}$$

where $y_i$ is the prediction of the model $M_i$ For decision-level aggregation, the weighted consensus decision is in Equation 20:

$$D_{BMA} = \sum w_i\, D_i \tag{20}$$

where $D_i$ denotes the decision from the model $M_i$ , and weights satisfy $\sum_i w_i = 1$. The normalized weights are defined as in Equation 21:

$$w_i = \frac{P(M_i|D)}{\sum_j P(M_j|D)} \tag{21}$$

Where the probability $P(M_i|D)$ is the posterior probability of model $M_i$ given the observed data $D$ , and the denominator represents the sum of posterior probabilities across all candidate models. BMA integrates model predictions according to their posterior probabilities, reducing reliance on any single predictor and improving robustness under workload variability [62,63]. In this study, posterior model weights are approximated using validation performance as a proxy for model evidence and are normalized via a softmax transformation to obtain probabilistic weights.

# 3. Related Work

This section situates our work within serverless autoscaling research. We organize prior studies by decision scope (placement vs. scaling), decision mechanism (rules, control, or learning), and use of structural information. We then highlight recurring limitations that motivate our framework.

## 3.1 Placement vs. Scaling Decisions

Several studies focus on execution placement, determining where function invocations execute. Weighted scheduling and multi-objective optimization improve throughput, latency, and energy efficiency by assigning requests to suitable nodes, particularly in heterogeneous or edge environments [64,65]. Platform-level designs such as OpenWhisk, OpenFaaS, and Fission introduce schedulers, admission control, and monitoring to enhance dispatch efficiency and reduce local queuing [66–68]. These efforts primarily optimize where requests execute rather than how much capacity should be allocated to each function. They do not explicitly address scaling decisions when requests traverse multiple dependent functions along a call graph. Our work complements placement strategies by directing scaling and pre-warming toward structurally critical functions that bound end-to-end performance.

## 3.1. Cold-Start Mitigation

Cold starts remain a major source of latency under bursty demand and scale-to-zero behavior. Stochastic optimization has been used to determine pre-spawning policies that balance responsiveness and energy cost [69]. However, many pre-warming approaches treat functions independently and do not exploit the dependency structure to prioritize which functions should be warmed first. In contrast, our framework leverages a dependency graph to identify bottleneck functions and target scaling and pre-warming accordingly.

## 3.2. Rule-Based and Control-Theoretic Approaches

Rule-based autoscaling remains widely adopted due to its simplicity and low overhead. Dynamic thresholds, escalation policies, mixed-resource rules, and deadline-aware schemes improve responsiveness relative to static policies [70–72]. Nevertheless, such approaches are typically reactive, sensitive to parameter tuning, and applied independently to each function. Control-theoretic and queueing-based methods regulate delay, utilization, or queue length under bursty workloads [42,73–75]. While these approaches offer stability guarantees, they generally operate at the service tier and rarely incorporate invocation-graph structure when allocating capacity. As a result, scaling decisions may not prioritize functions that dominate end-to-end latency.

### 3.3. Learning-Based Autoscaling

Learning-based methods enable proactive scaling through demand forecasting or policy learning. Reinforcement learning has been applied to elasticity control in systems such as Knative under non-stationary workloads [76–78]. Supervised predictors, including LSTM, CNN, and MLP models, forecast workload or resource utilization to reduce service violations under dynamic traces. Additional approaches employ profit-oriented provisioning or Markov models to estimate required instance counts [79–82]. Although these methods improve proactive decision-making, many rely on a single predictor and primarily use local signals. This can reduce robustness under workload drift, noise, and short-lived bursts. Hybrid controllers combining rules, prediction, and control have also been proposed [83–89], yet explicit dependency-aware prioritization and structured cold-start reasoning are often absent. Recent work explores multi-model forecasting to improve prediction accuracy. For example, [90] proposes an MLP–LSTM hybrid, [91] combines SVM, random forest, and deep neural networks, and investigates ensemble learning with hyperparameter tuning. Approaches such as [92] and [93] integrate CNN–LSTM architectures and vector autoregression for enhanced time-series prediction. While these models report improved forecasting accuracy, they typically focus on prediction alone and do not integrate scaling control, dependency-aware bottleneck targeting, or probabilistic consensus for model fusion.

### 3.4. Recurring Limitations

Across the literature, two limitations recur. First, inter-function dependencies are frequently ignored or handled indirectly, even though a small subset of functions along critical invocation paths can dominate tail latency and cost. Second, scaling decisions often rely on a single predictive model, reducing robustness under workload uncertainty and drift. The proposed framework addresses these gaps by modeling function interactions as a directed call graph, identifying bottleneck functions using graph-based measures, forecasting near-term demand with complementary lightweight models MLP, CNN, and LSTM, and combining predictions through Bayesian-inspired model averaging. This produces a dependency-aware, model-robust autoscaler that focuses scaling actions on functions with the greatest end-to-end impact. Table 1 summarizes the differences between our approach and representative learning-based methods.

**Table 1**. Comparison with Related Learning-Based Methods.

| Method | Architecture summary | Inputs | Preprocessing | Cold start handling | Limitations |
|---|---|---|---|---|---|
| **MLP-LSTM [90]** | Hybrid MLP + LSTM | CPU, memory, bandwidth | Normalization; windowing | No | Designed for VNF/SFC; no cold-start handling |
| **ML Models [91]** | SVM / RF / DNN | Big data metrics | Cleaning, normalization, and feature selection | No | Not serverless-aware; no cold-start handling |
| **ML with Hyperparameter Tuning [94]** | Ensemble ML on the cloud | Large-scale datasets | Normalization; hyperparameter tuning | No | Focuses on prediction only; no runtime autoscaling |
| **CNN-LSTM [92]** | CNN + LSTM | CPU, memory | Stationarity check, windowing | No | No autoscaling; single dataset |
| **DCNN LSTM [93]** | Deep CNN + LSTM | Resource-usage time series | Normalization; fixed budget | No | No autoscaling, Lower predictive accuracy |
| **Our Ensemble Method** | MLP + CNN + LSTM with consensus | CPU + platform signals | Normalization; sliding windows | Yes | Dynamic Autoscaling, better prediction accuracy |

# 4. The Proposed Approach

This section presents the proposed dependency-aware autoscaling framework for serverless applications, illustrated in Figure 3. The framework integrates structural analysis, runtime monitoring, predictive modeling, and consensus-based decision logic into a unified control pipeline. Specifically, dependency analysis constructs a watch set of high-impact functions; resource modeling generates runtime signals and action labels; supervised models forecast next-step demand for the watch set; the consensus module fuses predictions; and the cost-comparison module filters the proposed action before execution. The pipeline begins with Bottleneck Analysis, where the function dependency graph is analyzed using degree centrality to identify structurally critical functions. A watch set is constructed to prioritize monitoring and scaling decisions for these bottleneck candidates. In the Resource Modeling stage, runtime signals including CPU utilization, memory usage, and execution time are continuously monitored. Percentile-based adaptive thresholds generate baseline scaling indicators and detect high or low resource pressure. Next, Scalability Modeling predicts near-term resource demand for the watch set. Supervised learning models, LSTM, MLP, and CNN, provide complementary forecasts, while unsupervised techniques, SOM, K-means, and PCA, are included solely for comparative evaluation. Predictions from supervised models are aggregated in the Consensus module. Model performance weighting and Bayesian-inspired model averaging produce a robust consensus estimate. At this stage, cost comparison and cold-start awareness are incorporated to stabilize scaling decisions. Finally, the Scaling Decision module issues one of three actions: scale up, scale down, or hold, based on the consensus output. By combining dependency-aware prioritization, multi-model forecasting, probabilistic aggregation, and cost comparison, the framework enables targeted and stable horizontal scaling under dynamic workloads.

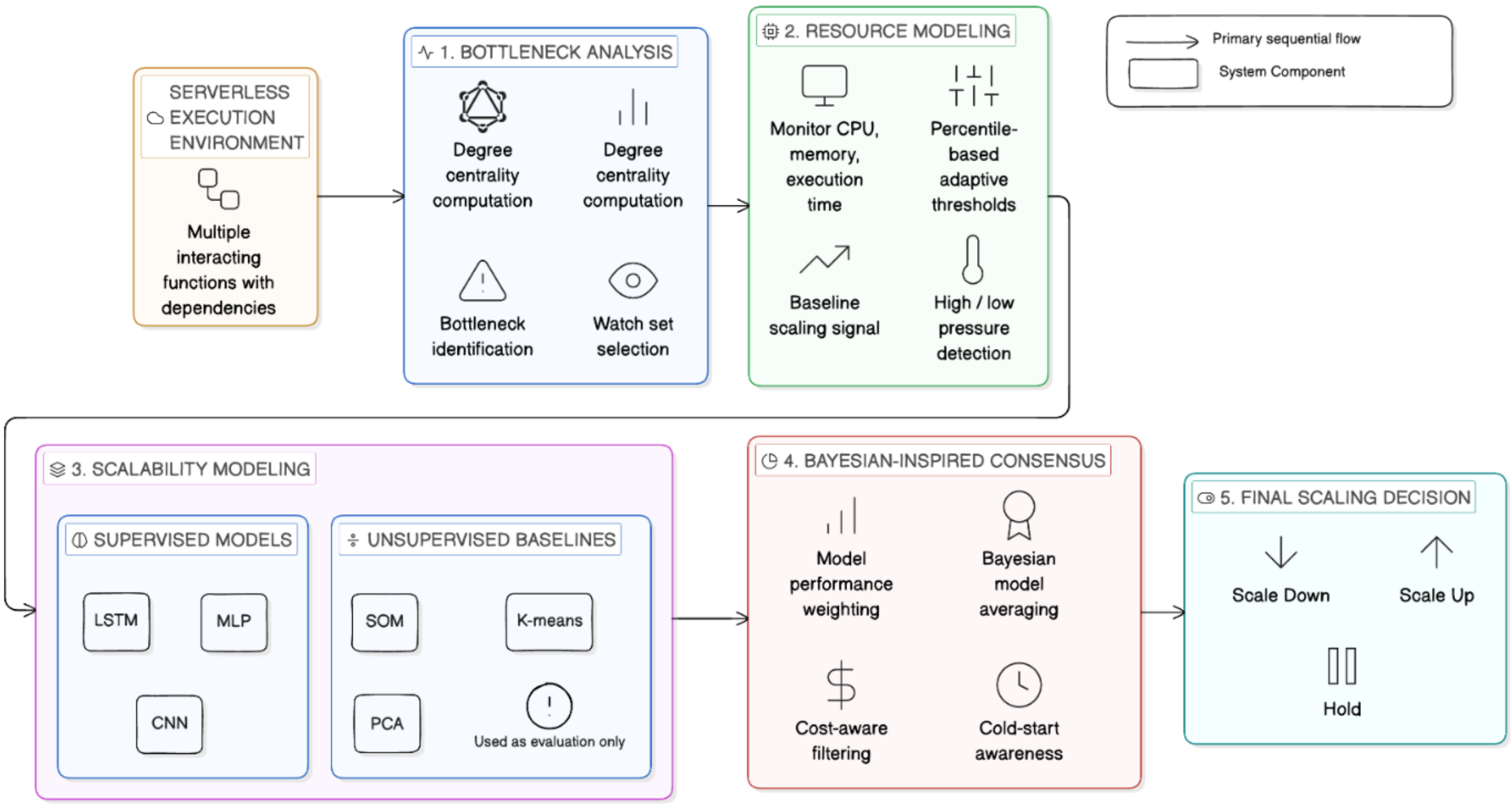

**Figure 3**. Dependency-aware multi-expert autoscaling framework architecture, illustrating bottleneck analysis, resource monitoring, predictive modeling, consensus, and final scaling decision.

## 4.1. Bottleneck Analysis Component

Serverless workflows typically consist of multiple interacting functions connected through invocation dependencies. Congestion in a single function can propagate along critical execution paths, significantly affecting end-to-end latency and cost. Rather than detecting overload reactively based solely on runtime metrics, the proposed framework proactively identifies structurally influential functions using dependency analysis. The application is represented as a directed dependency graph, where nodes correspond to functions and edges represent invocation relationships. Degree centrality is computed for each node to quantify its structural influence. This measure is selected for its low computational overhead and suitability for dynamic environments that require frequent updates. Functions with high degree centrality interact with many other components and are therefore more likely to amplify workload pressure across execution paths. The framework constructs a watch set consisting of the highest-ranked nodes. Subsequent monitoring, forecasting, and scaling decisions are prioritized for this reduced set of structurally critical functions.

### 4.2. Degree Centrality on the Dependency Graph

Bottleneck identification begins by constructing a weighted, directed dependency graph from the dependency dataset $D$ with columns $\{class, file, fanin, fanout\}$. Each distinct class and file is mapped to a unique node identifier, enabling consistent representation of dependency relationships. A directed edge is created from a class node $c_i$ to a file node $f_j$ whenever the class depends on the file. The edge weight captures dependency intensity using fan-in and fan-out metrics as defined in Equation 22:

$$w(c_i, f_j) = \text{fanin}(c_i) + \text{fanout}(c_i) \tag{22}$$

If multiple records exist for the same class–file pair, weights are accumulated to reflect repeated interactions. The resulting graph is defined as $G = (V, E, w)$, where the vertex set $V$ is partitioned into class nodes $V_C$, and file nodes $V_F$, and edges restricted to $E \subseteq V_C \times V_F$. This directed bipartite structure prevents artificial dependencies between nodes of the same type. For each class node $c_i \in V_C$, weighted degree centrality is defined as in Equation 23.

$$C_D(c_i) = \sum_{f_j \in N(c_i)} w(c_i, f_j) \tag{23}$$

where $N(c_i) \subseteq V_F$ denotes the set of file nodes connected to $c_i$. To enable ranking and comparison, centrality scores are normalized as in Equation 24.

$$C_D^{\text{norm}}(c_i) = \frac{C_D(c_i)}{\max_{c_k \in V_C} C_D(c_k)} \tag{24}$$

The bottleneck score is defined directly as the normalized degree centrality as defined in Equation 25.

$$\text{Score}(c_i) = C_D^{\text{norm}}(c_i) \tag{25}$$

Nodes are ranked in descending order of $\text{Score}(c_i)$, and the top-k nodes form the bottleneck watch set. Algorithm 1 summarizes this procedure and outputs the structurally high-impact functions that are subsequently prioritized for monitoring, forecasting, and scaling.

**Algorithm 1** Bottleneck Identification via Weighted Degree Centrality

**Input:** A directed weighted dependency graph $G = (V_C \cup V_F, E, w)$, where $V_C$ is the set of class nodes, $V_F$ is the set of file nodes, $E \subseteq V_C \times V_F$ represents dependency edges, and $w(c, f)$ denotes the weight of dependency $(c, f)$; and an integer $k$, the number of bottlenecks to return.
**Output:** Top-$k$ bottleneck class nodes ranked by bottleneck score.
**Compute weighted degree centrality**
**for** each class node $c \in V_C$ **do**
  $C_D(c) \leftarrow 0$
  **for** each edge $(c, f) \in E$ **do**
    $C_D(c) \leftarrow C_D(c) + w(c, f)$
  **end for**
**end for**
**Normalize degree centrality**
$C_{\max} \leftarrow \max_{c \in V_C} C_D(c)$
**for** each class node $c \in V_C$ **do**
  **if** $C_{\max} > 0$ **then**
    $C_D^{\text{norm}}(c) \leftarrow \frac{C_D(c)}{C_{\max}}$
  **else**
    $C_D^{\text{norm}}(c) \leftarrow 0$
  **end if**
**end for**
**Rank bottlenecks**
Sort class nodes $c \in V_C$ in descending order of $C_D^{\text{norm}}(c)$
**return** top-$k$ class nodes

### 4.3. Resource Modeling Component

After identifying bottleneck functions, the resource modeling component derives adaptive scaling thresholds from historical runtime behavior to support stable control decisions. Let $D_{\text{hist}}$ denote the historical time series of runtime metrics for a selected bottleneck. Rather than relying on fixed thresholds, the controller computes percentile-based bounds that reflect empirical workload behavior. The component continuously monitors execution time ($\text{Runs_ms} \in \mathbb{R}^+$), average CPU utilization ($\text{AvgCPU} \in [0,100]$), per-core CPU utilization ($\text{CPUCore}[j] \in [0,100]$), memory usage ($\text{MemUsed} \in \mathbb{R}^+$), and available memory ($\text{MemAvail} \in \mathbb{R}^+$). Let $P_p(X)$ denote the $p$-th percentile of metric $X$ computed from $D_{\text{hist}}$. Upper thresholds, such as $P_{95}$ capture high-pressure conditions that may require scaling up, while intermediate and lower thresholds, such as $P_{70}$ and $P_{05}$, define safe operating region. The use of distinct upper and lower bounds introduces hysteresis and reduces oscillatory behavior. Scaling decisions follow a prioritized three-way rule. A high-pressure state is triggered if any monitored metric exceeds its upper threshold. Logical OR is used to ensure responsiveness to emerging bottlenecks. If no high-pressure condition is detected, a low-pressure state is evaluated. Scale-down occurs only when all monitored metrics fall within safe bounds. The Logical AND operator is used to prevent premature downscaling. If neither condition is satisfied, the controller maintains the current resource allocation. Algorithm 2 summarizes the complete threshold derivation and decision process.

**Algorithm 2** Resource Modeling and Scaling Decision Using Historical Thresholds

**Input:** historical metrics $D_{\text{hist}}$; current metrics $\mathbf{x}$ with fields Run_ms, AvgCPU, CPUCore[1..4], MemUsed, MemAvail
**Output:** decision $\in \{-1, 0, 1\}$ ($1$ = SCALE_UP, $0$ = HOLD, $-1$ = SCALE_DOWN) and thresholds
**Step 1: Derive thresholds from history**
$\tau_{cpu}^{hi} \leftarrow P_{95}(D_{\text{hist}}.\text{AvgCPU})$
$\tau_{cpu}^{lo} \leftarrow P_{70}(D_{\text{hist}}.\text{AvgCPU})$
$\tau_{mem}^{hi} \leftarrow P_{95}(D_{\text{hist}}.\text{MemUsed})$
$\tau_{mem}^{lo} \leftarrow P_{70}(D_{\text{hist}}.\text{MemUsed})$
$\tau_{avail}^{lo} \leftarrow P_{05}(D_{\text{hist}}.\text{MemAvail})$
$\tau_{run}^{hi} \leftarrow P_{95}(D_{\text{hist}}.\text{Run_ms})$
thresholds $\leftarrow \{\tau_{cpu}^{hi}, \tau_{cpu}^{lo}, \tau_{mem}^{hi}, \tau_{mem}^{lo}, \tau_{avail}^{lo}, \tau_{run}^{hi}\}$
**Aggregate current metrics**
peakCPU $\leftarrow \max_{j \in \{1,2,3,4\}} \mathbf{x}.\text{CPUCore}[j]$
**Step 3: Evaluate high-pressure condition**
cpuHigh $\leftarrow (\mathbf{x}.\text{AvgCPU} > \tau_{cpu}^{hi}) \lor (\text{peakCPU} > \tau_{cpu}^{hi})$
memHigh $\leftarrow (\mathbf{x}.\text{MemUsed} > \tau_{mem}^{hi})$
availLow $\leftarrow (\mathbf{x}.\text{MemAvail} < \tau_{avail}^{lo})$
runHigh $\leftarrow (\mathbf{x}.\text{Run_ms} > \tau_{run}^{hi})$
pressure$_{hi}$ $\leftarrow$ cpuHigh $\lor$ memHigh $\lor$ availLow $\lor$ runHigh
**if** pressure$_{hi}$ **then**
    **return** $(1, \text{thresholds})$ ▷ SCALE_UP
**end if**
**Step 4: Evaluate low-pressure condition**
cpuLow $\leftarrow (\mathbf{x}.\text{AvgCPU} < \tau_{cpu}^{lo}) \land (\text{peakCPU} < \tau_{cpu}^{lo})$
memLow $\leftarrow (\mathbf{x}.\text{MemUsed} < \tau_{mem}^{lo})$
availSafe $\leftarrow (\mathbf{x}.\text{MemAvail} > \tau_{avail}^{lo})$
pressure$_{lo}$ $\leftarrow$ cpuLow $\land$ memLow $\land$ availSafe
**if** pressure$_{lo}$ **then**
    **return** $(-1, \text{thresholds})$ ▷ SCALE_DOWN
**else**
    **return** $(0, \text{thresholds})$ ▷ HOLD
**end if**

## 4.4. Scalability Component

The scalability component produces short-horizon demand predictions that enable proactive scaling decisions. To evaluate learning paradigms under a consistent decision framework, both unsupervised and supervised methods are implemented and compared. However, only supervised models are used to drive the final autoscaling decisions. Unsupervised methods identify latent structure in normalized runtime metrics without access to scaling labels. We consider three clustering strategies: K-Means, PCA followed by K-Means, and self-organizing maps. Let $X$ denote the normalized feature matrix constructed from runtime metrics. Each method produces cluster assignments $z \in \{1, \ldots, k\}$. Because cluster indices do not directly correspond to scaling actions, a post-processing step maps each cluster to one of the three scaling decisions $\{-1, 0, 1\}$ (scale-down, hold, scale-up). The mapping is defined using majority voting on labeled training data: each cluster is assigned the most frequent scaling label among its members. This mapping remains fixed during evaluation to ensure comparability with supervised methods. Algorithm 3 details the full procedure.

**Algorithm 3** Unsupervised Learning-Based Scalability Decision (K-Means, PCA+K-Means, SOM)

**Input:** dataset $D$; number of clusters $k$ (default 3); SOM grid shape $(h, w)$; PCA components $p$; random seed seed
**Output:** cluster labels $z_{\text{km}}, z_{\text{pca}}, z_{\text{som}}$
**Step 1: Build feature matrix**
$X \leftarrow \textsc{ExtractFeatures}(D)$ ▷ $X$ uses $\{\text{Run_ms}, \text{AvgCPU}, \text{CPUCore}[1..4], \text{MemUsed}, \text{MemAvail}\}$
**Step 2: Normalize features**
$X_s \leftarrow \textsc{MinMaxScale}(X, [0, 1])$
**Step 3: K-Means clustering on normalized features**
$\text{km} \leftarrow \textsc{KMeans}(k, \text{seed})$
$\text{km.fit}(X_s)$
$z_{\text{km}} \leftarrow \text{km.predict}(X_s)$
**Step 4: SOM-based clustering**
$\text{som} \leftarrow \text{SOM}(\text{grid} = (h, w),\ \text{dim} = \text{cols}(X_s),\ \sigma = 1.0,\ \eta = 0.5,\ \text{seed})$
$\text{som.init_weights}(X_s)$
$\text{som.train}(X_s, \text{iters} = 100)$
$W \leftarrow [\,\text{som.winner}(x) \text{ for each } x \in X_s\,]$ ▷ $W$ are winner grid coordinates
$z_{\text{som}} \leftarrow \textsc{KMeans}(k, \text{seed}).\textsc{fit_predict}(W)$
**Step 5: PCA + K-Means**
$\text{pca} \leftarrow \text{PCA}(p, \text{seed})$
$X_p \leftarrow \text{pca.fit_transform}(X_s)$
$\text{km}_p \leftarrow \textsc{KMeans}(k, \text{seed})$
$z_{\text{pca}} \leftarrow \text{km}_p.\textsc{fit_predict}(X_p)$
**return** $z_{\text{km}}, z_{\text{pca}}, z_{\text{som}}$

In contrast, supervised learning directly models the mapping from runtime metrics to scaling actions. The problem is formulated as a three-class classification task with inputs derived from CPU utilization, memory usage, and execution time features. We evaluate three lightweight neural architectures selected for complementary modeling capabilities while maintaining low inference overhead: The MLP model captures nonlinear relationships among aggregated features, the CNN extracts localized patterns within sliding windows of resource metrics, and the LSTM captures temporal dependencies across short look-back windows. Time-series samples are constructed using a fixed look-back window $\ell$. Features are standardized prior to training. Each model outputs class probabilities through a softmax layer. Model configurations are intentionally compact to preserve online feasibility. The MLP uses one hidden layer with 64 units and ReLU activation, followed by dropout (0.2) and a softmax output. The LSTM model uses a single LSTM layer with 64 units, dropout (0.2), and softmax output; to limit inference latency, inputs are provided as a short sequence window. The CNN applies a one-dimensional convolution with 32 filters of width three, global max pooling, a dense layer with 64 units, and a softmax output. All models are trained using the Adam optimizer with categorical cross-entropy loss and early stopping to prevent overfitting. Time-series cross-validation is applied to preserve temporal structure. Algorithm 4 summarizes the supervised training procedure. Performance is evaluated with accuracy, precision, recall, and F1-score, as well as Mean Absolute Error (MAE), Mean Squared Error (MSE), averaged across time-series cross-validation folds.

**Algorithm 4** Supervised Learning-Based Scalability Decision

**Input:** dataset $D$; labels $y$; look-back window $\ell$; number of folds $k$; early-stopping patience $p$
**Output:** trained models $M_{\text{mlp}}, M_{\text{lstm}}, M_{\text{cnn}}$; performance summary $R$
**Step 1: Feature preprocessing**
$X \leftarrow \textsc{ExtractFeatures}(D)$
scaler $\leftarrow \textsc{StandardScaler}().\textsc{fit}(X)$
$X_s \leftarrow$ scaler.$\textsc{transform}(X)$
**Step 2: Create time-series samples**
$(X_{\text{seq}}, y_{\text{seq}}) \leftarrow \textsc{CreateLaggedDataset}(X_s, y, \ell)$
$y_{\text{oh}} \leftarrow \textsc{OneHotEncode}(y_{\text{seq}})$
**Step 3: Time-series cross-validation**
$\text{CV} \leftarrow \textsc{TimeSeriesSplit}(k)$
$R \leftarrow \emptyset$
**for** each $(\text{train}, \text{val}) \in \text{CV}$ **do**
  $X_{tr}, X_{val} \leftarrow X_{\text{seq}}[\text{train}], X_{\text{seq}}[\text{val}]$
  $y_{tr}, y_{val} \leftarrow y_{\text{oh}}[\text{train}], y_{\text{oh}}[\text{val}]$
  **MLP model**
  $X^f_{tr} \leftarrow \textsc{Flatten}(X_{tr})$
  $X^f_{val} \leftarrow \textsc{Flatten}(X_{val})$
  $M_{\text{mlp}} \leftarrow \textsc{BuildMLP}()$
  $\hat{y}_{\text{mlp}} \leftarrow \textsc{TrainEval}(M_{\text{mlp}}, X^f_{tr}, y_{tr}, X^f_{val}, y_{val}, p)$
  $R \leftarrow R \cup \textsc{Metrics}(\hat{y}_{\text{mlp}}, y_{val}, \text{MLP})$
  **LSTM model**
  $M_{\text{lstm}} \leftarrow \textsc{BuildLSTM}()$
  $\hat{y}_{\text{lstm}} \leftarrow \textsc{TrainEval}(M_{\text{lstm}}, X_{tr}, y_{tr}, X_{val}, y_{val}, p)$
  $R \leftarrow R \cup \textsc{Metrics}(\hat{y}_{\text{lstm}}, y_{val}, \text{LSTM})$
  **CNN model**
  $M_{\text{cnn}} \leftarrow \textsc{BuildCNN}()$
  $\hat{y}_{\text{cnn}} \leftarrow \textsc{TrainEval}(M_{\text{cnn}}, X_{tr}, y_{tr}, X_{val}, y_{val}, p)$
  $R \leftarrow R \cup \textsc{Metrics}(\hat{y}_{\text{cnn}}, y_{val}, \text{CNN})$
**end for**
$R \leftarrow \textsc{AggregateByModel}(R)$
**return** $M_{\text{mlp}}, M_{\text{lstm}}, M_{\text{cnn}}, R$

### 4.5. Consensus component

Relying on a single predictor can make scaling decisions sensitive to workload drift, noise, or transient bursts. The consensus component addresses this by combining predictions from the MLP, CNN, and LSTM using a performance-weighted probabilistic ensemble inspired by Bayesian model averaging principles. Let $M = \{M_{MLP}, M_{LSTM}, M_{CNN}\}$ denote the trained models. Each model produces class probability estimates $P_i$ on the test data. Model performance is evaluated on validation data to obtain accuracy scores $A_i$ that reflect recent predictive reliability. Model weights are computed using a softmax transformation with temperature parameter τ as defined in Equation 26.

$$\mathrm{w_i} = \frac{\exp(\frac{A_i}{\tau})}{\sum_j \exp(\frac{A_j}{\tau})} \tag{26}$$

This formulation ensures higher-performing models receive greater influence, All weights remain positive and $\sum_i w_i = 1$. The ensemble probability distribution is then computed as in Equation 27.

$$P_{ens} = \sum_i w_i P_i \tag{27}$$

Final scaling decisions are obtained via Equation 28.

$$\hat{y}_{ens} = \arg\max_c P_{ens}(c) \tag{28}$$

This performance-weighted probabilistic aggregation reduces dependence on any single model while preserving complementary predictive information. Algorithm 5 details the complete consensus procedure.

**Algorithm 5** Consensus Component via Performance-Weighted Ensemble

**Input:** trained models $M = \{M_{\text{MLP}}, M_{\text{LSTM}}, M_{\text{CNN}}\}$; validation data $(X_{\text{val}}, y_{\text{val}})$; test data $X_{\text{test}}$; temperature $\tau$
**Output:** ensemble predictions $\hat{y}_{\text{ens}}$; model weights $w$; ensemble probabilities $P_{\text{ens}}$
**Step 1: Evaluate model performance**
**for** each model $M_i \in M$ **do**
  $A_i \leftarrow \text{EVALUATE}(M_i, X_{\text{val}}, y_{\text{val}})$
  $P_i \leftarrow \text{PREDICT}(M_i, X_{\text{test}})$
**end for**
**Step 2: Compute model weights**
$w_i \leftarrow \exp(A_i/\tau)$ for all $i$
$w \leftarrow w / \sum_i w_i$
**Step 3: Aggregate predictions**
$P_{\text{ens}} \leftarrow \sum_i w_i \cdot P_i$
**Step 4: Generate final decisions**
**for** each test sample $k$ **do**
  $\hat{y}_{\text{ens}}[k] \leftarrow \arg\max_c P_{\text{ens}}[k, c]$
**end for**
**return** $\hat{y}_{\text{ens}}, w, P_{\text{ens}}$

### 4.6. Cost Comparison Component

The cost comparison component ensures that consensus-driven scaling actions remain economically justified by estimating their projected operational impact before execution. While overprovisioning increases direct resource expenditure, underprovisioning increases latency and may lead to SLA violations; therefore, scaling decisions must balance cost efficiency and performance reliability. Operational cost at time t is modeled as the sum of three components: base resource cost, waste cost, and overload cost. The base cost represents pay-per-use charges proportional to the number of allocated replicas. Waste cost captures overprovisioning when allocated capacity exceeds actual utilization, reflecting underutilized resources. Overload cost models underprovisioning penalties when demand exceeds capacity, and this is amplified when SLA violations occur. In particular, SLA violations are detected through latency thresholds, and the overload penalty is increased accordingly. This mechanism implicitly incorporates cold-start effects, since scale-down decisions that increase predicted latency beyond SLA thresholds raise the overload cost, discouraging aggressive deprovisioning. Algorithm 6 formalizes the computation of these cost terms and derives the projected total cost.

**Algorithm 6** Cost Component: Cost Calculation

**Input:** current replicas $R_t$; actual utilization $U_t$; predicted utilization $\hat{U}_t$; actual latency $L_t$; SLA threshold $L_{\text{SLA}}$; unit cost $C_{\text{base}}$; multipliers $C_{\text{waste}}, C_{\text{overload}}$; maximum replicas $R_{\max}$
**Output:** $C_{\text{total}}(t)$; $C_{\text{base}}(t)$; $C_{\text{waste}}(t)$; $C_{\text{overload}}(t)$; cost_savings$(t)$
**Step 1: Base resource cost**
$C_{\text{base}}(t) \leftarrow R_t \cdot C_{\text{base}}$
**Step 2: Waste cost (over-provisioning)**
$W_t \leftarrow \max(0, \hat{U}_t - U_t)$
$C_{\text{waste}}(t) \leftarrow R_t \cdot C_{\text{base}} \cdot W_t \cdot C_{\text{waste}}$
**Step 3: Overload cost (under-provisioning)**
$O_t \leftarrow \max(0, U_t - \hat{U}_t)$
$P_t \leftarrow \mathbb{I}[L_t > L_{\text{SLA}}]$ ▷ $P_t = 1$ if SLA violated, else 0
$C_{\text{overload}}(t) \leftarrow R_t \cdot C_{\text{base}} \cdot O_t \cdot C_{\text{overload}} \cdot (1 + P_t)$
**Step 4: Total cost and savings (vs. baseline)**
$C_{\text{total}}(t) \leftarrow C_{\text{base}}(t) + C_{\text{waste}}(t) + C_{\text{overload}}(t)$
$C_{\text{baseline}}(t) \leftarrow R_{\max} \cdot C_{\text{base}}$
cost_savings$(t) \leftarrow C_{\text{baseline}}(t) - C_{\text{total}}(t)$
**return** $C_{\text{total}}(t), C_{\text{base}}(t), C_{\text{waste}}(t), C_{\text{overload}}(t)$, cost_savings$(t)$

Given a consensus scaling decision, the controller proposes a candidate replica configuration subject to replica bounds. The projected cost of the candidate configuration is then estimated using predicted next-step utilization and latency. Algorithm 7 performs cost-aware acceptance or rejection using a tolerance margin δ. If the candidate cost is sufficiently lower than the current cost, the scaling action is accepted; if it exceeds the allowable tolerance, the action is rejected, and the system holds the current allocation. This comparison-based filtering prevents economically unjustified scaling, particularly scale-down actions that could trigger cold-start latency or SLA degradation. To ensure runtime feasibility, the overall framework remains computationally lightweight. Degree centrality is computed in linear time with respect to the number of dependency edges and can be updated incrementally. The supervised models are intentionally shallow, and inference requires only a small number of matrix operations per control interval. Retraining is performed periodically using a sliding window of recent observations rather than at every step, limiting overhead while preserving adaptability.

**Algorithm 7** Cost-Aware Scaling Decision

**Input:** consensus decision $\hat{y}_{\text{cons}} \in \{\textsc{ScaleDown}, \textsc{Hold}, \textsc{ScaleUp}\}$; current replicas $R_t$; current cost $C_{\text{total}}(t)$; predicted next-step utilization $\hat{U}_{t+1}$; predicted next-step latency $\hat{L}_{t+1}$; cost tolerance $\delta$; replica bounds $R_{\min}, R_{\max}$
**Output:** next replicas $R_{t+1}$; scaling action action
**Step 1: Propose candidate replicas from consensus**
**if** $\hat{y}_{\text{cons}} = \textsc{ScaleUp}$ **then**
    $R_{\text{cand}} \leftarrow R_t + 1$
**else if** $\hat{y}_{\text{cons}} = \textsc{ScaleDown}$ **then**
    $R_{\text{cand}} \leftarrow R_t - 1$
**else**
    $R_{\text{cand}} \leftarrow R_t$
**end if**
$R_{\text{cand}} \leftarrow \min\big(R_{\max}, \max(R_{\min}, R_{\text{cand}})\big)$ ▷ Clamp to bounds
**Step 2: Estimate candidate cost**
$C_{\text{cand}} \leftarrow \textsc{EstimateCost}(R_{\text{cand}}, \hat{U}_{t+1}, \hat{L}_{t+1})$
**Step 3: Cost-aware acceptance/rejection**
**if** $C_{\text{cand}} < (1-\delta)\, C_{\text{total}}(t)$ **then**
    $R_{t+1} \leftarrow R_{\text{cand}}$
    action $\leftarrow \hat{y}_{\text{cons}}$
**else if** $C_{\text{cand}} > (1+\delta)\, C_{\text{total}}(t)$ **then**
    $R_{t+1} \leftarrow R_t$
    action $\leftarrow \textsc{Hold}$
**else**
    $R_{t+1} \leftarrow R_{\text{cand}}$
    action $\leftarrow \hat{y}_{\text{cons}}$
**end if**
**return** $R_{t+1}$, action

## 5. Evaluations

The proposed framework is evaluated to assess prediction accuracy, decision stability, and cost impact under dynamic workload conditions. Section 5.1 describes the experimental environment. Section 5.2 details the datasets used for dependency modeling and resource-consumption forecasting. Subsequent sections analyze bottleneck identification, threshold-based scaling behavior, unsupervised baselines, supervised prediction performance, consensus aggregation, and cost-aware decision outcomes.

### 5.1. Environment and Implementation

The framework is implemented in Python. TensorFlow and Keras are used to train the supervised learning models, and Pandas supports preprocessing and feature construction. Experiments are conducted in the Google Colab environment.
The implementation follows the modular architecture described in Section 4, including dependency analysis, percentile-based thresholding, supervised forecasting, performance-weighted ensemble aggregation, and cost-aware decision filtering.

## 5.2. Dataset Preparation

Two datasets are used in this study: one for dependency-driven bottleneck identification and one for resource-consumption forecasting. To approximate structural coupling among serverless functions, we use a publicly available software dependency dataset containing code-level structural metrics from a large software project. Although this dataset is not collected from a deployed serverless platform, it serves as a reproducible and structured proxy for modeling dependency relationships. In the absence of publicly available serverless invocation-graph traces, this dataset enables controlled evaluation of structurally informed prioritization strategies while maintaining transparency and reproducibility independent of proprietary platform data. The dataset contains 509,426 records and 53 attributes, where each record corresponds to a file or class component. From this dataset, we extract the File identifier, Class identifier, Fan-in (number of incoming dependencies), Fan-out (number of outgoing dependencies), and Total connections (fan-in + fan-out). These attributes are used to construct a weighted directed bipartite dependency graph, where classes and files form distinct node sets and dependency relations define directed edges. Degree centrality is then computed to identify structurally influential components. Table 2 summarizes the extracted attributes used in the analysis.

**Table 2**. Overview of the dependency analysis dataset.

| | |
|---|---|
| **File** | Name of the source file |
| **Class** | Name of the class |
| **Fan-in** | Number of incoming dependencies |
| **Fan-out** | Number of outgoing dependencies |
| **Total-con** | fan-in + fan-out |

For workload modeling, resource consumption data are obtained from real hardware performance traces collected using OpenHardwareMonitor [12]. The dataset consists of time-series measurements of CPU utilization, per-core CPU usage, memory utilization, and execution time. These traces serve as workload inputs for demand forecasting and autoscaling evaluation. Before analysis, both datasets undergo cleaning and normalization. The dependency dataset is used exclusively to compute centrality scores and identify structurally high-impact nodes. To evaluate bottleneck-aware scaling under realistic workload dynamics, we select representative bottleneck targets based on structural ranking and treat each selected target as the monitored function whose resource time series drives forecasting and scaling decisions. This design isolates the structural prioritization mechanism from workload generation, allowing the dependency-aware control logic to be evaluated using real performance traces without requiring proprietary serverless call-graph data.

**Table 3**. Sample Records from the Dependency Analysis Dataset

| file | class | Fan-in | Fan-out | Total-con |
|---|---|---|---|---|
| **\DATASET\abfiles\r0\c17\after\ActiveFiltersPanelTest.java** | mtgdeckbuilder.frontend.ActiveFiltersPanelTest | 0 | 4 | 4 |
| **\DATASET\abfiles\r0\c17\after\TagViewerTest.java** | mtgdeckbuilder.frontend.TagViewerTest | 1 | 6 | 7 |
| **\DATASET\abfiles\r0\c17\after\TagTopicTest.java** | mtgdeckbuilder.frontend.topics.TagTopicTest | 0 | 4 | 4 |
| **\DATASET\abfiles\r0\c17\after\TagAddPanel.java** | mtgdeckbuilder.frontend.TagAddPanel$Anonymous1 | 0 | 1 | 1 |
| **\DATASET\abfiles\r0\c17\after\TagTopic.java** | mtgdeckbuilder.frontend.topics.TagTopic | 12 | 1 | 13 |
| **\DATASET\abfiles\r0\c17\after\NewFilterPanelTest.java** | mtgdeckbuilder.frontend.NewFilterPanelTest | 0 | 4 | 4 |
| **\DATASET\abfiles\r0\c17\after\TagAddPanel.java** | mtgdeckbuilder.frontend.TagAddPanel | 3 | 4 | 7 |
| **\DATASET\abfiles\r0\c17\after\CardTaggingPanelTest.java** | mtgdeckbuilder.frontend.CardTaggingPanelTest | 1 | 8 | 9 |

## 5.3. Bottleneck Identification and Threshold-Based Decisions

The dependency graph constructed from the structural dataset is analyzed to identify high-impact nodes using weighted degree centrality. Figure 4 presents the top bottleneck candidates ranked by normalized centrality score. The most connected node exhibits 224 total dependency connections, followed by nodes with 196 and 137 connections. These nodes interact with a large number of components and therefore are structurally positioned to influence multiple execution paths. In the evaluation pipeline, the highest-ranked node is selected as the bottleneck watch target. Its associated resource time series is used for subsequent forecasting and scaling experiments. To generate baseline scaling decisions, runtime metrics of the selected bottleneck are compared against percentile-based thresholds derived from historical behavior. Upper thresholds correspond to the 95th percentile for CPU utilization, memory usage, and execution time, while available memory uses the 5th percentile to detect critically low capacity. Lower thresholds are defined at the 70th percentile for CPU and memory utilization and the 10th percentile for available memory, introducing hysteresis to reduce oscillatory behavior. Scaling actions are produced using the three-way logic defined in Section 4.3: scale-up is triggered if any metric exceeds its upper bound, scale-down occurs only when all metrics fall below their lower bounds, and hold is issued otherwise. Grounding thresholds in empirical distributions reduces sensitivity to transient fluctuations while preserving responsiveness to sustained load increases. In addition to serving as a baseline controller, this threshold mechanism provides consistent action labels for supervised model training and for mapping unsupervised clusters to discrete scaling decisions.

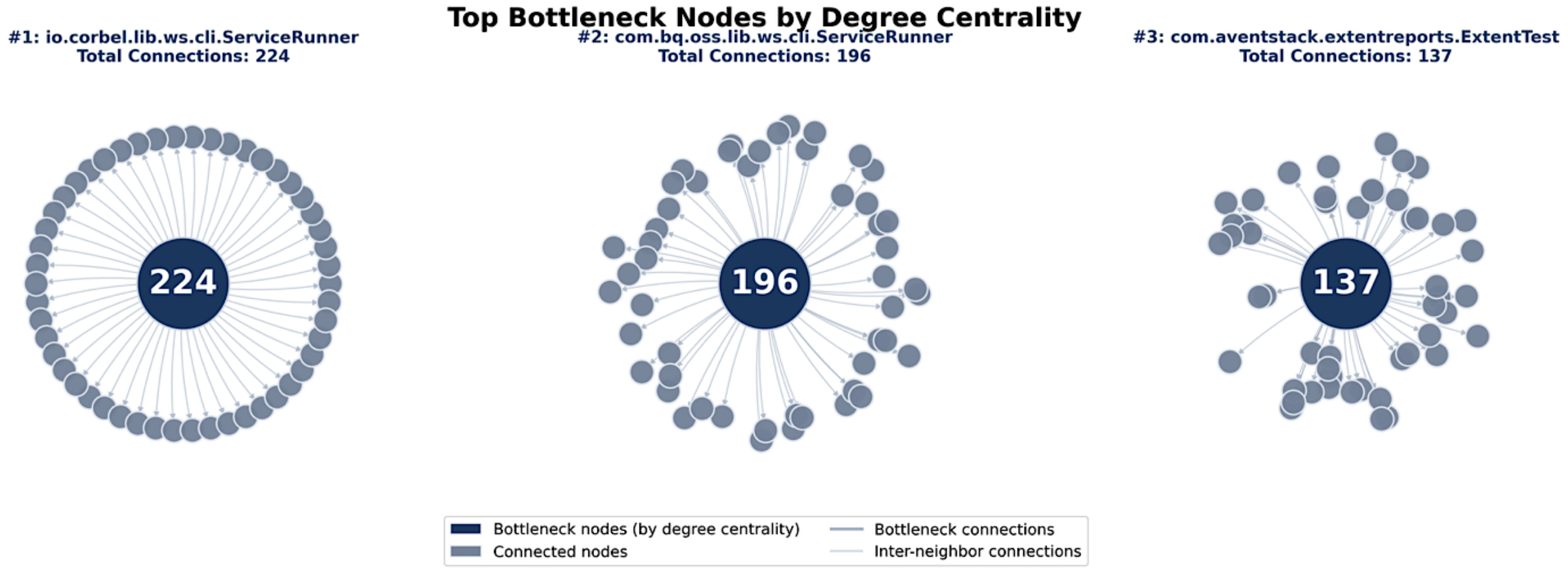

**Figure 4**. Bottleneck candidates ranked by normalized degree centrality

## 5.4. Unsupervised Learning Evaluations

We evaluate K-Means, Self-Organizing Maps (SOM), and PCA-based clustering to assess whether unlabeled grouping of resource traces can recover scaling decisions without supervision. Cluster assignments are post-processed using majority voting to map each cluster to one of the three scaling actions (scale-down, hold, scale-up), ensuring comparability with supervised classifiers. Figure 5 presents K-Means clustering visualized using t-SNE. While the projection reveals partially distinguishable groups, the clusters do not align consistently with scaling labels, resulting in an accuracy of 49.5%. Figure 6 shows SOM-based clustering. The resulting clusters exhibit substantial overlap between scaling categories, and decision alignment accuracy drops to 21.6%. Figure 7 presents K-Means clustering applied after PCA dimensionality reduction. The projection fails to meaningfully separate scaling behaviors, producing near-random performance with an accuracy of 0.6%. These results demonstrate that unsupervised pattern discovery, even when visually separable in reduced-dimensional space, does not reliably recover the control logic required for autoscaling. Structural grouping alone is insufficient to capture the directional decision boundaries needed for scale-up and scale-down actions.

.

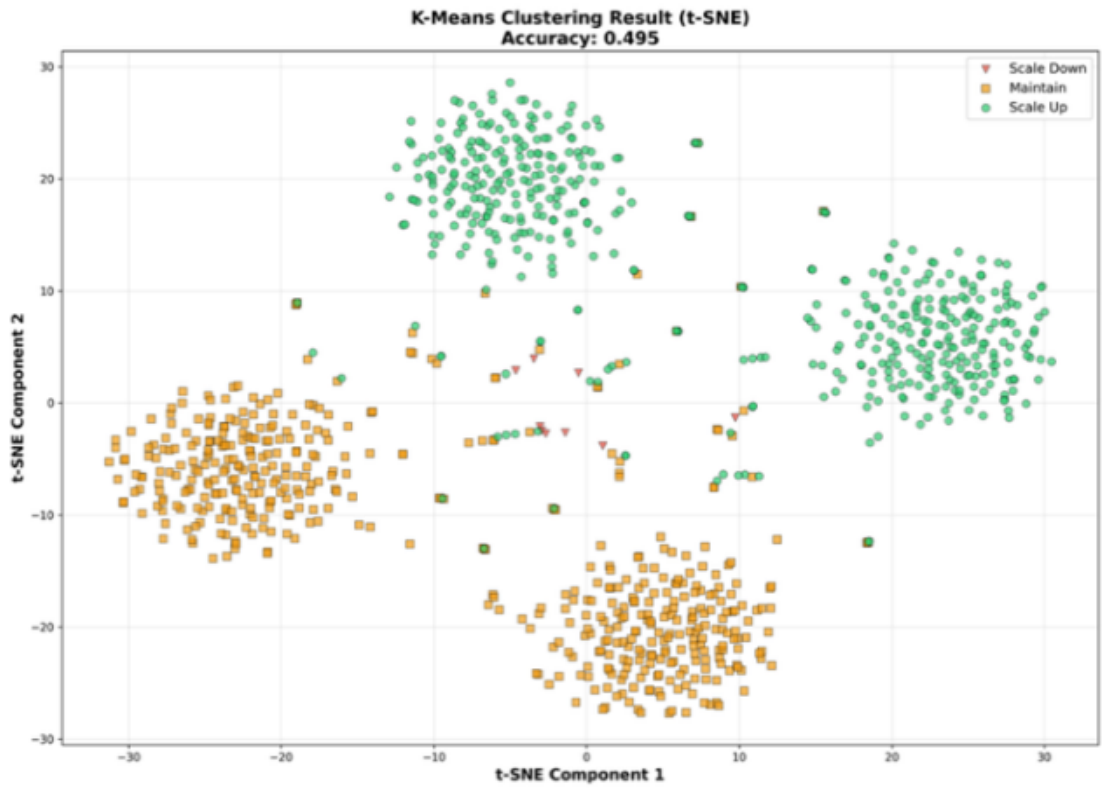

**Figure 5**. K-Means Clustering Results Visualized with t-SNE

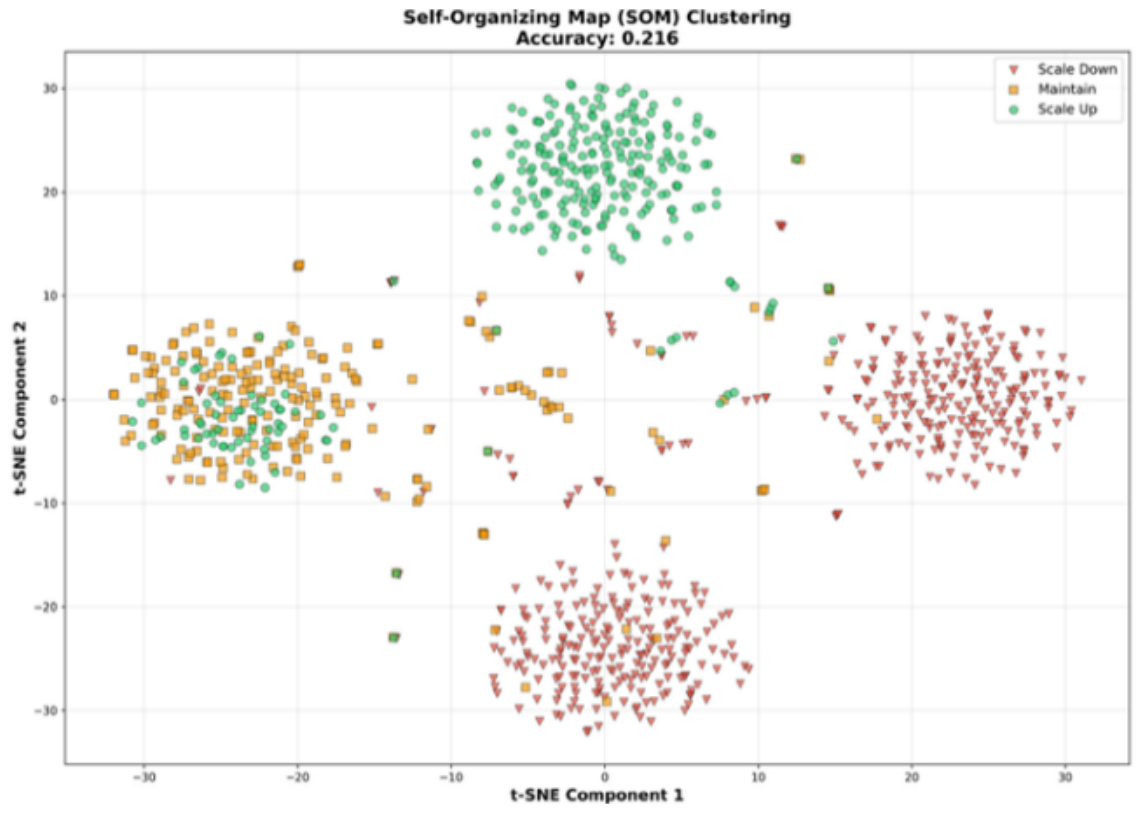

**Figure 6.** Self-Organizing Map (SOM) Clustering Visualized with t-SNE

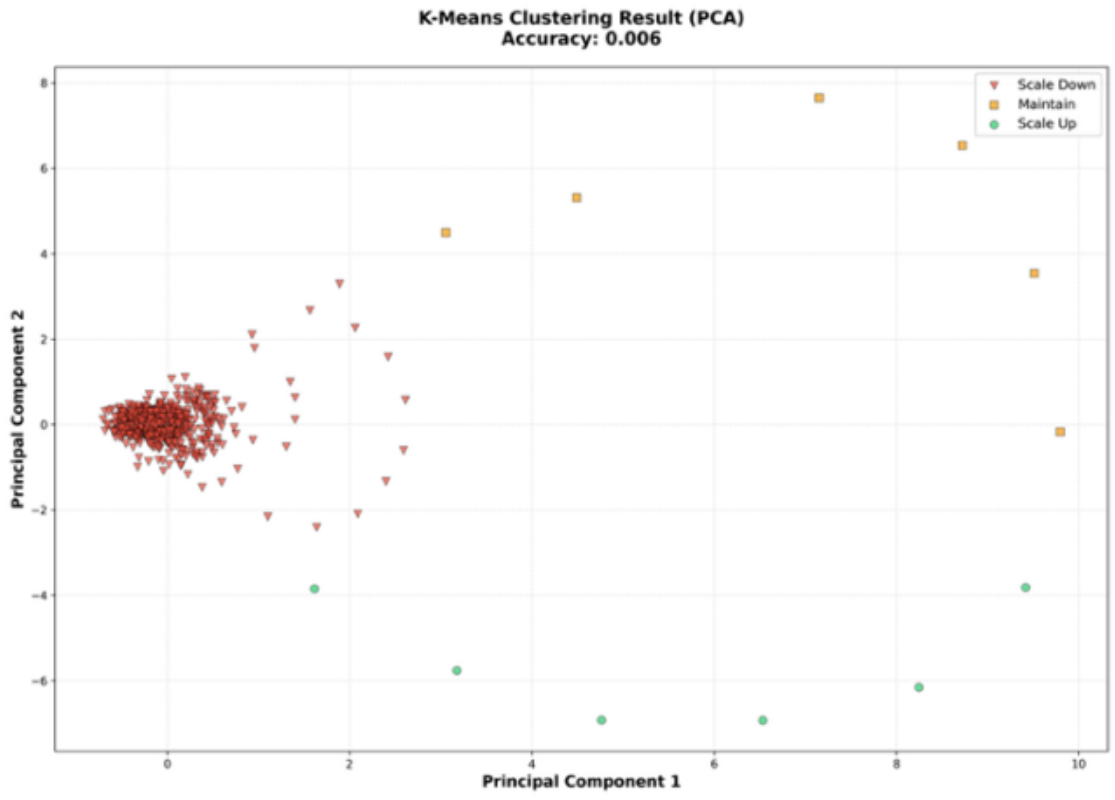

**Figure 7**. K-Means Clustering Results Visualized with PCA

Table 4 compares unsupervised clustering approaches with supervised neural models (MLP, LSTM, and CNN) under identical input features. The supervised models achieve consistently high predictive performance. The LSTM obtains the highest average accuracy (99.06%), followed by MLP (98.75%) and CNN (98.50%). Precision, recall, and F1-scores exceed 0.98 across models, indicating stable classification across scaling categories. Error metrics (MAE and MSE) are correspondingly low.

In contrast, the strongest unsupervised baseline (K-Means with t-SNE visualization) achieves only 49.5% accuracy, while SOM and PCA-based clustering perform substantially worse. The large performance gap of nearly 50 percentage points indicates that labeled supervision is critical for accurate autoscaling decision generation. These findings confirm that clustering-based pattern discovery cannot substitute for predictive modeling in time-sensitive scaling control. Consequently, the final controller relies on supervised forecasting combined with consensus aggregation rather than unsupervised grouping.

**Table 4**. Comparative performance analysis of supervised and unsupervised approaches for autoscaling decision prediction.

| Approach | Model | MAE | MSE | Precision | Recall | F1-Score | Accuracy (%) |
|---|---|---|---|---|---|---|---|
| **Supervised** | MLP | 0.02 | 0.02 | 0.98 | 0.99 | 0.98 | 98.75 |
| | LSTM | 0.01 | 0.01 | 0.98 | 0.99 | 0.99 | 99.06 |
| | CNN | 0.02 | 0.04 | 0.98 | 0.99 | 0.98 | 98.50 |
| **Unsupervised** | K-Means (t-SNE) | - | - | - | - | - | 49.5 |
| | SOM | - | - | - | - | - | 21.6 |
| | K-Means (PCA) | - | - | - | - | - | 0.6 |

### 5.5. Supervised Learning Evaluations

In the supervised phase, autoscaling is formulated as a three-class classification problem using labels generated by the adaptive threshold mechanism described in Section 4.3. The dataset is divided into 80% training, 10% validation, and 10% for testing. To assess generalization, five-fold cross-validation is performed. Model performance is evaluated using accuracy, precision, recall, F1-score, MAE, MSE, and Root Mean Squared Error (RMSE). Figure 8 represents the training and validation curves for the MLP, LSTM, and CNN models over 100 epochs. All three architectures converge rapidly during early training and stabilize thereafter. Training and validation curves remain closely aligned across accuracy, MAE, and MSE, indicating minimal overfitting. Among the models, the LSTM exhibits the most stable convergence with consistently smooth validation curves. The MLP shows slightly greater variability, while the CNN maintains stable behavior but with marginally higher final error values. These learning dynamics indicate that all models are well-regularized and suitable for real-time inference.

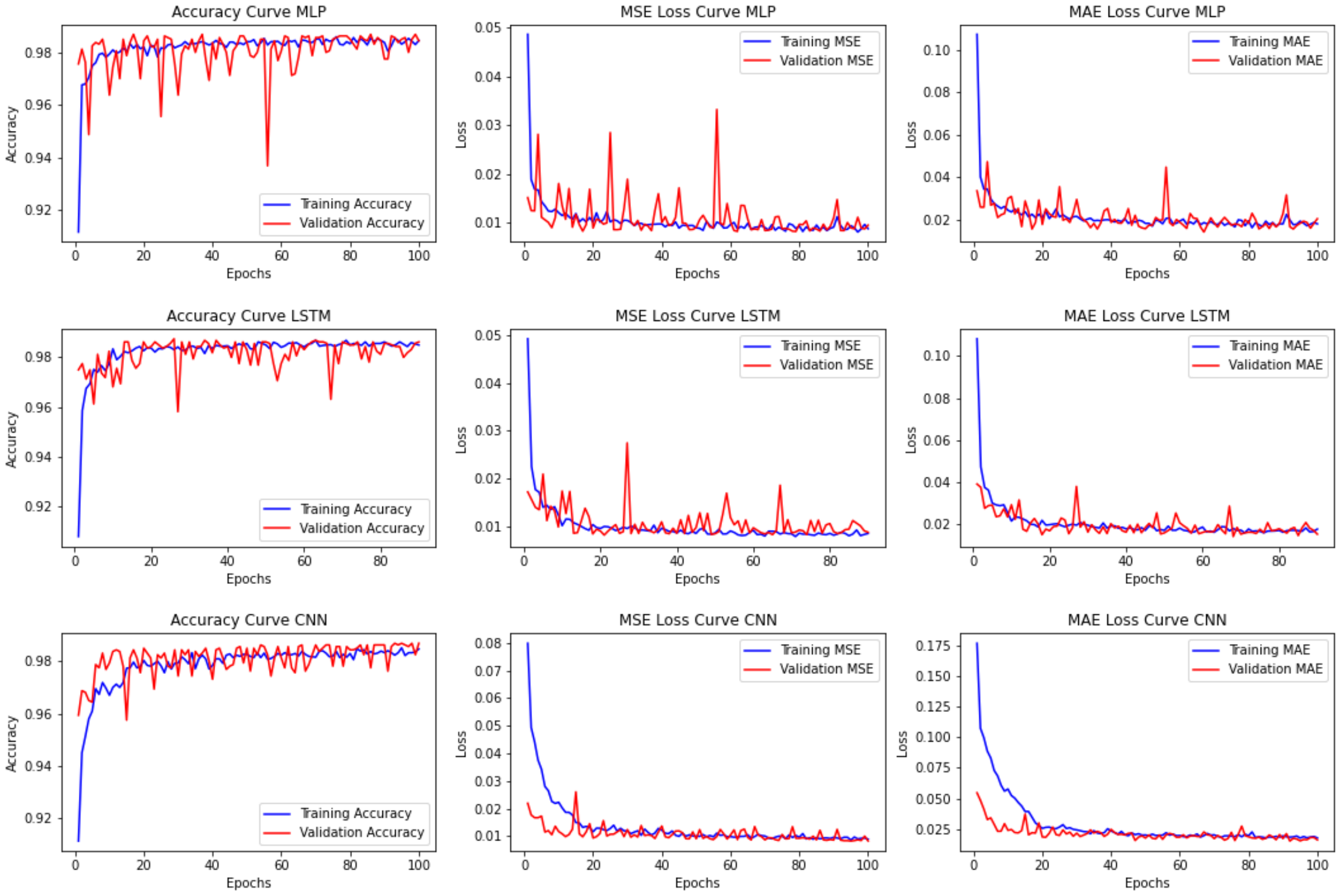

**Figure 8**. Training and validation performance of the supervised learning models (MLP – first row, LSTM – second row, CNN – third row) over 100 epochs. The columns report classification accuracy, MSE, and MAE. All models demonstrate stable convergence with closely aligned training and validation curves, with the LSTM exhibiting the most consistent and robust performance.

Figure 9 compares final error metrics across models. The LSTM achieves the lowest MAE, MSE, and RMSE values, confirming its strong temporal modeling capability. The MLP demonstrates competitive performance with slightly higher error values. The CNN performs well overall but exhibits higher variability in certain folds. The five-fold cross-validation results in Table 5, Table 6, and Table 7 further support these findings. The LSTM consistently achieves the lowest error and highest stability across folds. Although all supervised models outperform unsupervised baselines by a large margin, the LSTM provides the strongest standalone predictor.

**Table 5.** Five-fold cross-validation performance of the MLP model, including MAE, MSE, F1-score, recall, precision, and accuracy across folds and averaged results.

| | **Performance Analysis of MLP** | | | | **Error Analysis of MLP** | |
|---|---|---|---|---|---|---|
| | **Accuracy** | **Precision** | **Recall** | **F1-Score** | **MSE** | **MAE** |
| **Fold 1** | 98.40% | 0.98 | 0.98 | 0.98 | 0.04 | **0.02** |
| **Fold 2** | 99.00% | 0.98 | 0.99 | 0.99 | 0.02 | **0.01** |
| **Fold 3** | 98.75% | 0.98 | 0.99 | 0.98 | 0.03 | **0.02** |
| **Fold 4** | 98.90% | 0.98 | 0.99 | 0.99 | 0.02 | **0.01** |
| **Fold 5** | 98.70% | 0.98 | 0.99 | 0.98 | 0.02 | **0.02** |
| **Average** | **98.75%** | **0.98** | **0.99** | **0.98** | **0.02** | **0.02** |

**Table 6.** Five-fold cross-validation performance of the LSTM model, including MAE, MSE, F1-score, recall, precision, and accuracy across folds and averaged results.

| | **Performance Analysis of LSTM** | | | | **Error Analysis of LSTM** | |
|---|---|---|---|---|---|---|
| | **Accuracy** | **Precision** | **Recall** | **F1-Score** | **MSE** | **MAE** |
| **Fold 1** | 99.20% | 0.99 | 0.99 | 0.99 | 0.01 | **0.01** |
| **Fold 2** | 98.95% | 0.98 | 0.99 | 0.99 | 0.02 | **0.01** |
| **Fold 3** | 99.05% | 0.98 | 0.99 | 0.99 | 0.01 | **0.01** |
| **Fold 4** | 99.05% | 0.98 | 0.99 | 0.99 | 0.01 | **0.01** |
| **Fold 5** | 99.05% | 0.98 | 0.99 | 0.99 | 0.01 | **0.01** |
| **Average** | **99.06%** | **0.98** | **0.99** | **0.99** | 0.01 | **0.01** |

**Table 7**. Five-fold cross-validation performance of the CNN model, including MAE, MSE, F1-score, recall, precision, and accuracy across folds and averaged results.

| | **Performance Analysis of CNN** | | | | **Error Analysis of CNN** | |
|---|---|---|---|---|---|---|
| | **Accuracy** | **Precision** | **Recall** | **F1-Score** | **MSE** | **MAE** |
| **Fold 1** | 98.75% | 0.98 | 0.99 | 0.98 | 0.02 | **0.02** |
| **Fold 2** | 98.65% | 0.98 | 0.99 | 0.98 | 0.03 | **0.02** |
| **Fold 3** | 97.15% | 0.97 | 0.97 | 0.97 | 0.09 | **0.05** |
| **Fold 4** | 99.00% | 0.98 | 0.99 | 0.99 | 0.02 | **0.01** |
| **Fold 5** | 98.95% | 0.98 | 0.99 | 0.99 | 0.02 | **0.01** |
| **Average** | **98.5%** | **0.98** | **0.99** | **0.98** | **0.04** | **0.02** |

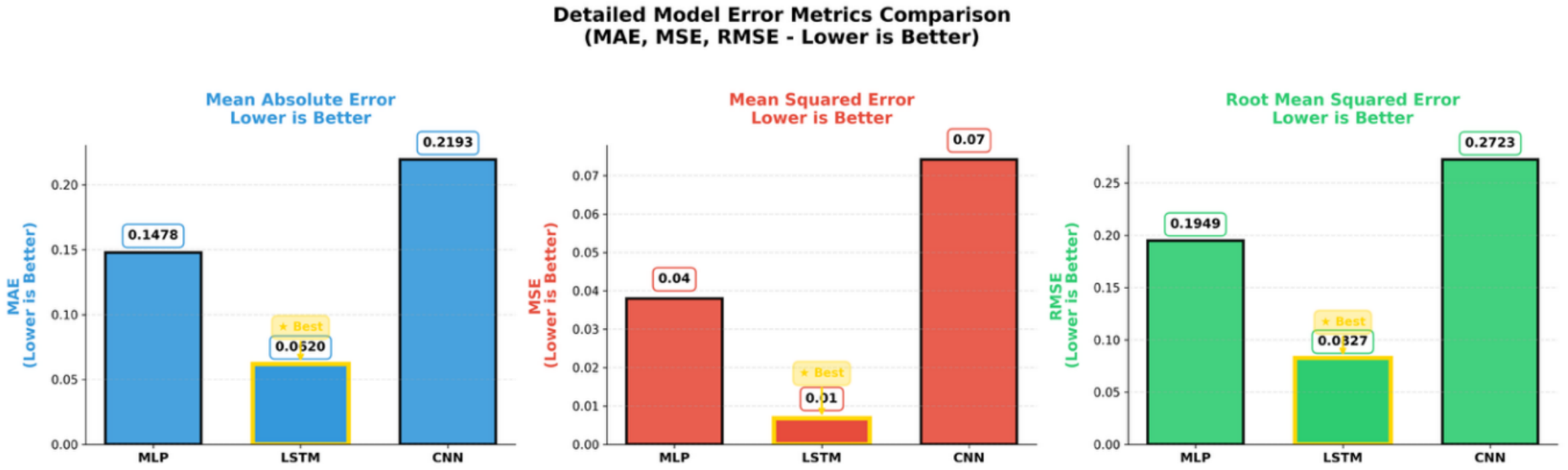

**Figure 9**. Comparative evaluation of prediction error metrics for the MLP, LSTM, and CNN supervised models. The models are assessed using MAE, MSE, and RMSE, where lower values indicate better performance.

Table 8 and Table 9 compare individual model predictions with the performance-weighted probabilistic ensemble for representative samples at time t and one-step-ahead forecasts t+1. While individual predictors occasionally deviate from actual values, the ensemble consistently reduces extreme deviations by weighting model outputs according to validation performance. When a single model underestimates or overshoots resource demand, the aggregated prediction moderates the error through probabilistic averaging. This behavior demonstrates that consensus aggregation reduces reliance on any single model and improves stability under workload variability. The ensemble prediction tracks actual resource values more consistently than individual models, supporting the design choice of performance-weighted fusion in the final controller.

**Table 8**. Comparison of actual and predicted resource values at the current time step (t) using MLP, LSTM, CNN, and Bayesian-inspired model averaging.

| **Bayesian-inspired Avg Prediction for t** | **CNN Prediction for t** | **LSTM Prediction for t** | **MLP Prediction for t** | **Actual Value for t** |
|---|---|---|---|---|
| 9.934422 | 9.263559 | 10.364485 | 10.236974 | **10.3520393** |
| 5.103557 | 4.7756 | 5.405163 | 5.160522 | **5.2734375** |
| 29.304556 | 29.035881 | 29.725983 | 29.177689 | **29.3650742** |

**Table 9**. Comparison of actual and predicted resource values at the next time step (t+1) using MLP, LSTM, CNN, and Bayesian-inspired model averaging.

| **Bayesian-inspired Avg Prediction for t+1** | **CNN Prediction for t+1** | **LSTM Prediction for t+1** | **MLP Prediction for t+1** | **Actual Value for t+1** |
|---|---|---|---|---|
| 59.63367 | 46.664474 | 64.61939 | 63.02586 | **64.19699** |
| 59.659576 | 46.92247 | 64.57348 | 62.944954 | **64.19621** |
| 59.811844 | 47.328598 | 64.62794 | 63.03141 | **64.20816** |

### 5.6. Comparison with Related Forecasting Methods

To position the proposed framework within the current literature, we compare it against recent neural forecasting architectures, including hybrid and ensemble approaches [90–94]. All methods are evaluated using the same workload traces and performance metrics to ensure a fair comparison. Unlike hybrid architectures that merge multiple networks into a single composite predictor (e.g., CNN–LSTM or MLP–LSTM), the proposed framework maintains model diversity by training independent predictors and aggregating their outputs using a performance-weighted probabilistic ensemble. This design preserves complementary signal representations while reducing sensitivity to noise, local minima, and workload drift. Table 10 summarizes the comparative results. The proposed approach achieves the highest overall accuracy and the lowest MAE, MSE, and RMSE among all evaluated methods. In comparison, the closest competing hybrid model (CNN–LSTM [92]) achieves 99.85% accuracy with a higher MSE (0.0060). Deep CNN–LSTM [93] reports 99.62% accuracy and higher error values across all metrics. The MLP–LSTM hybrid [90] achieves 99.49% accuracy but exhibits nearly four times higher MSE than the proposed approach. Traditional machine learning models [91] achieve substantially lower accuracy (75.03%) and significantly higher error rates. The proposed framework improves accuracy by up to 4.85 percentage points over traditional ML baselines and achieves substantial reductions in MSE relative to hybrid deep models. In addition to predictive accuracy, the proposed method explicitly incorporates cold-start awareness within its cost-aware control logic, whereas the compared forecasting models focus solely on prediction performance without integrating scaling control or SLA-aware filtering. These results demonstrate that separating model expertise and reconciling predictions via performance-weighted aggregation yields both improved quantitative accuracy and greater robustness than monolithic hybrid architectures.

**Table 10.** Performance comparison of the proposed ensemble approach with representative baseline methods. The proposed method achieves superior predictive accuracy and error reduction while explicitly addressing cold-start effects.

| Method | Accuracy | MAE | MSE | RMSE | Cold-Start Aware |
|---|---|---|---|---|---|
| **Proposed Ensemble Approach** | **0.9988** | **0.0021** | **0.0039** | **0.0403** | **Yes** |
| MLP-LSTM [90] | 0.9949 | 0.0090 | 0.0168 | 0.1263 | No |
| ML Models [91] | 0.7503 | 0.4946 | 0.9843 | 0.9871 | No |
| ML with Hyperparameter Tuning [94] | 0.9978 | 0.0031 | 0.0049 | 0.0483 | No |
| CNN-LSTM [92] | 0.9985 | 0.0030 | 0.0060 | 0.0565 | No |
| Deep CNN-LSTM [93] | 0.9962 | 0.0060 | 0.0104 | 0.0906 | No |

### 5.7. Autoscaler Cost Impact

We evaluate the economic impact of the proposed controller while maintaining response latency near the target SLA. The controller forecasts near-term demand at one-minute intervals and converts predictions into discrete scaling actions. A cooldown mechanism limits oscillatory behavior, and scale-down decisions are issued only when predicted demand remains low and latency stays within a guard band. Scale-to-zero is applied selectively based on projected economic benefit and anticipated cold-start impact. At each decision step, the total cost is computed using platform-specific pricing models. For function-based services, cost includes request charges and memory-time billing, whereas container-based services use vCPU- and memory-second pricing. The cost formulation incorporates overload penalties and SLA violation indicators. Cold-start effects are modeled indirectly: scale-down actions that increase predicted latency beyond the SLA threshold incur additional penalty, reflecting the expected delay when insufficient warm capacity is available. This penalty discourages aggressive deprovisioning that could otherwise produce economically unfavorable latency spikes.

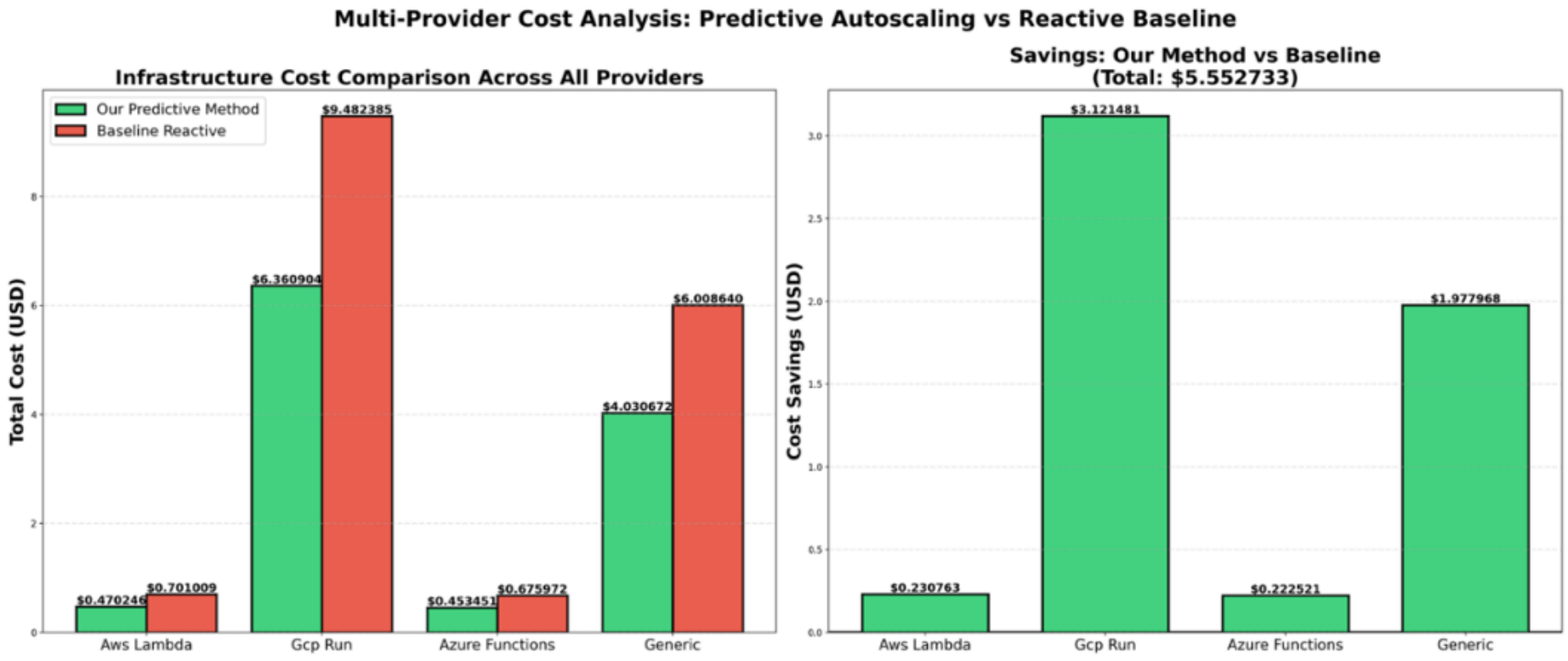

**Figure 10**. Total infrastructure cost and corresponding savings achieved by the proposed predictive autoscaling method.

Figure 10 reports total infrastructure cost and corresponding savings relative to a reactive threshold-based baseline. Across all evaluated platforms, the predictive controller consistently reduces total cost while preserving performance targets. On AWS Lambda, the cost decreases from $0.70 to $0.47. On Google Cloud Run, the cost decreases from $9.48 to $6.36. On Azure Functions, the cost decreases from $0.68 to $0.45. Under a generic container pricing model, the cost decreases from $6.01 to $4.03. The aggregated savings across platforms amount to $5.55 relative to the reactive baseline. These savings result from two complementary effects. First, proactive scale-up mitigates prolonged latency excursions that would otherwise trigger delayed and excessive reactive scaling. Second, controlled scale-down reduces idle replica time while avoiding SLA violations and cold-start penalties. Together, these mechanisms demonstrate that dependency-aware predictive autoscaling can achieve economically efficient resource allocation without compromising responsiveness or stability.

# 6. Conclusion

This paper presents a dependency-aware autoscaling framework for serverless environments that addresses structural, predictive, and economic limitations of existing approaches. By modeling function interactions as a directed dependency graph and identifying structurally influential bottlenecks using degree centrality, the framework prioritizes scaling actions on components with the greatest end-to-end impact. Near-term resource demand is predicted using lightweight supervised models (MLP, LSTM, and CNN), and their outputs are reconciled through a performance-weighted probabilistic ensemble to improve stability and reduce sensitivity to model-specific bias. Our approach incorporates cold-start awareness and cost-comparison logic to balance latency guarantees with operational efficiency. Experimental results demonstrate that supervised learning substantially outperforms unsupervised clustering for autoscaling decision generation. The proposed ensemble approach achieves higher predictive accuracy and lower error than representative hybrid forecasting models while maintaining stable training dynamics. End-to-end experiments across multiple pricing models show consistent cost reductions, without compromising SLA targets. Overall, this work shows that integrating dependency-aware bottleneck targeting, supervised multi-model forecasting, and cost-aware control yields a robust and economically efficient solution for practical serverless autoscaling.

# 7. Future Work

Although the proposed framework advances dependency-aware and ensemble-based autoscaling, several directions remain for further improvement in adaptability, scalability, and interpretability. These extensions aim to evolve the framework from predictive autoscaling toward more adaptive and intelligent resource management in next-generation serverless systems.

1. Federated learning may enable collaborative model training across multiple serverless platforms without sharing raw workload data. This approach would allow platforms to benefit from shared knowledge while preserving privacy. Open challenges include handling workload heterogeneity, personalizing global models to local behavior, and ensuring secure and stable aggregation of distributed updates.

2. Deploying the proposed framework on real Function-as-a-Service platforms such as AWS Lambda or Knative would enable evaluation under fully operational conditions. Integrating dependency inference, workload forecasting, and actuation within live systems would support analysis of long-term stability, convergence behavior, and tail latency performance under bursty and non-stationary demand.
3. Quantum-inspired optimization offers a potential avenue for improving scalability in large dependency graphs. By formulating scaling decisions as multi-objective optimization problems, such heuristics may provide efficient approximations for jointly optimizing latency, cost, and resource utilization in complex systems.
4. Graph neural networks (GNNs) could extend the framework to dynamic dependency modeling. Unlike static graph metrics, temporal GNNs may learn evolving interaction patterns and detect emerging critical paths, enabling earlier bottleneck identification and more proactive resource allocation.
5. Incorporating causal inference into the autoscaling logic may enhance both robustness and interpretability. By identifying which factors directly cause latency or overload, rather than relying only on correlated signals, the controller can make more reliable scaling decisions when workload patterns change and provide clearer explanations for its actions.

# Ethical approval and consent to participate

This article does not contain any studies with human participants or animals performed by any of the authors.

# Availability of data and material

Derived data supporting the findings of this study are publicly available in a GitHub repository and can be shared upon request.

# Competing interests

All authors declare that they have no conflicts of interest.

# Funding

The authors did not receive support from any organization for the submitted work.

# References

[1] Tari M, Ghobaei-Arani M, Pouramini J. Auto-scaling mechanisms in serverless computing: A comprehensive review. Comput Sci Rev 2024;53. https://doi.org/10.1016/j.cosrev.2024.100650.

[2] Tournaire T, Castel-Taleb H, Hyon E. Efficient Computation of Optimal Thresholds in Cloud Auto-scaling Systems. ACM Trans Model Perform Eval Comput Syst 2023;8. https://doi.org/10.1145/3603532.

[3] Wen J, Chen Z, Jin X, Liu X. Rise of the Planet of Serverless Computing: A Systematic Review. ACM Transactions on Software Engineering and Methodology 2023;32:1–61. https://doi.org/10.1145/3579643.

[4] Jonas E, Schleier-Smith J, Sreekanti V, Tsai C-C, Khandelwal NJ, Pu Q, Venkataramani V, Carreira J, Krauth K, Yadwadkar N, Gonzalez JE, Popa RA, Stoica I, Patterson DA. Cloud Programming Simplified: A Berkeley View on Serverless Computing 2019.

[5] Bibal Benifa J V, Dejey D. RLPAS: Reinforcement Learning-based Proactive Auto-Scaler for Resource Provisioning in Cloud Environment. Mobile Networks and Applications 2019;24:1348–63. https://doi.org/10.1007/s11036-018-0996-0.

[6] Manner J. A Structured Literature Review Approach to Define Serverless Computing and Function as a Service. IEEE International Conference on Cloud Computing, CLOUD, vol. 2023- July, IEEE Computer Society; 2023, p. 516–22. https://doi.org/10.1109/CLOUD60044.2023.00068.

[7] Shafiei H, Khonsari A, Mousavi P. Serverless Computing: A Survey of Opportunities, Challenges, and Applications. ACM Comput Surv 2022;54:1–32. https://doi.org/10.1145/3510611.

[8] Song S, Tong H, Meng C, Pan M, Yu Y. FuncScaler: Cold-Start-Aware Holistic Autoscaling for Serverless Resource Management. Proceedings of the IEEE International Conference on Web Services, ICWS, Institute of Electrical and Electronics Engineers Inc.; 2024, p. 1036–47. https://doi.org/10.1109/ICWS62655.2024.00122.
[9] Xu L, Saxena D, Yadwadkar NJ, Akella A, Gupta I. Dirigo: Self-scaling Stateful Actors For Serverless Real-time Data Processing 2023.
[10] Wen J, Chen Z, Liu X. A Literature Review on Serverless Computing 2022.
[11] Mampage A, Karunasekera S, Buyya R. A Holistic View on Resource Management in Serverless Computing Environments: Taxonomy and Future Directions. ACM Comput Surv 2022;54. https://doi.org/10.1145/3510412.
[12] Madsen K vejrup. Performance Data 2022. https://doi.org/10.5281/zenodo.14262334.
[13] Li B, Xia T, Wang W, Zhang J, Wang S, Cao C, Dong Z. FluidEdge: Expediting Serverless Machine Learning Inference via Bottleneck-aware Auto-scaling on Edge SoCs. IEEE Trans Mob Comput 2025. https://doi.org/10.1109/TMC.2025.3592334.
[14] Ma~kiewicz A, Ratajczak W. PRINCIPAL COMPONENTS ANALYSIS (PCA)*. vol. 19. 1993.
[15] Abdi H, Williams LJ. Principal component analysis. Wiley Interdiscip Rev Comput Stat 2010;2:433–59. https://doi.org/10.1002/wics.101.
[16] Denton PB, Parke SJ, Tao T, Zhang X. Eigenvectors from eigenvalues: A survey of a basic identity in linear algebra 2021. https://doi.org/10.1090/bull/1722.
[17] Rokach L, Maimon O. Clustering Methods. Data Mining and Knowledge Discovery Handbook, Springer-Verlag; 2006, p. 321–52. https://doi.org/10.1007/0-387-25465-x_15.
[18] Berahmand K, Saberi-Movahed F, Sheikhpour R, Li Y, Jalili M. A Comprehensive Survey on Spectral Clustering with Graph Structure Learning 2025.
[19] Ling LS, Weiling CT. Enhancing Segmentation: A Comparative Study of Clustering Methods. IEEE Access 2025;13:47418–39. https://doi.org/10.1109/ACCESS.2025.3550339.
[20] Ahmed M, Seraj R, Islam SMS. The k-means algorithm: A comprehensive survey and performance evaluation. Electronics (Switzerland) 2020;9:1–12. https://doi.org/10.3390/electronics9081295.
[21] Abdelsamea MM, Gnecco G, Gaber MM. A Survey of SOM-Based Active Contour Models for Image Segmentation. Advances in Intelligent Systems and Computing, vol. 295, Springer Verlag; 2014, p. 293–302. https://doi.org/10.1007/978-3-319-07695-9_28.
[22] Qu X, Yang L, Guo K, Ma L, Sun M, Ke M, Li M. A Survey on the Development of Self-Organizing Maps for Unsupervised Intrusion Detection. Mobile Networks and Applications 2021;26:808–29. https://doi.org/10.1007/s11036-019-01353-0.
[23] Florida S, Liu Y, Weisberg RH. A Review of Self-Organizing Map Applications in Meteorology and A Review of Self-Organizing Map Applications in Meteorology and Oceanography Oceanography Scholar Commons Citation Scholar Commons Citation "A Review of Self-Organizing Map Applications in Meteorology 14 A Review of Self-Organizing Map Applications in Meteorology and Oceanography. n.d.
[24] Chang YI. A Survey: Potential Dimensionality Reduction Methods 2025.
[25] Mittal M, Praveen Gujjar J, Guru Prasad MS, Devadas RM, Ambreen L, Kumar V. Dimensionality Reduction using UMAP and TSNE Technique. 2nd IEEE International Conference on Advances in Information Technology, ICAIT 2024 - Proceedings, Institute of Electrical and Electronics Engineers Inc.; 2024. https://doi.org/10.1109/ICAIT61638.2024.10690797.
[26] Rana A, Singh Rawat A, Bijalwan A, Bahuguna H. Application of Multi Layer (Perceptron) Artificial Neural Network in the Diagnosis System: A Systematic Review. 2018 International Conference on Research in Intelligent and Computing in Engineering (RICE), IEEE; 2018, p. 1–6. https://doi.org/10.1109/RICE.2018.8509069.
[27] Popescu M-C, Balas VE. Multilayer Perceptron and Neural Networks. n.d. https://doi.org/10.5555/1639537.1639542.
[28] Singh J, Banerjee R. A Study on Single and Multi-layer Perceptron Neural Network. 2019 3rd International Conference on Computing Methodologies and Communication (ICCMC), IEEE; 2019, p. 35–40. https://doi.org/10.1109/ICCMC.2019.8819775.
[29] Smagulova K, James AP. A survey on LSTM memristive neural network architectures and applications. European Physical Journal: Special Topics 2019;228:2313–24. https://doi.org/10.1140/epjst/e2019-900046-x.
[30] Landi F, Baraldi L, Cornia M, Cucchiara R. Working Memory Connections for LSTM 2021. https://doi.org/10.1016/j.neunet.2021.08.030.
[31] Zhao J, Huang F, Lv J, Duan Y, Qin Z, Li G, Tian G. Do RNN and LSTM have Long Memory? 2020.

[32] Pulver A, Lyu S. LSTM with working memory. 2017 International Joint Conference on Neural Networks (IJCNN), IEEE; 2017, p. 845–51. https://doi.org/10.1109/IJCNN.2017.7965940.
[33] Yao K, Cohn T, Vylomova K, Duh K, Dyer C. Depth-Gated LSTM. n.d.
[34] Lu Y, Salem FM. Simplified Gating in Long Short-term Memory (LSTM) Recurrent Neural Networks. CoRR 2017;abs/1701.03441.
[35] Siami-Namini S, Tavakoli N, Namin AS. The Performance of LSTM and BiLSTM in Forecasting Time Series. 2019 IEEE International Conference on Big Data (Big Data), IEEE; 2019, p. 3285–92. https://doi.org/10.1109/BigData47090.2019.9005997.
[36] Alzubaidi L, Zhang J, Humaidi AJ, Al-Dujaili A, Duan Y, Al-Shamma O, Santamaría J, Fadhel MA, Al-Amidie M, Farhan L. Review of deep learning: concepts, CNN architectures, challenges, applications, future directions. J Big Data 2021;8. https://doi.org/10.1186/s40537-021-00444-8.
[37] Elngar AA, Arafa M, Fathy A, Moustafa B, Mahmoud O, Shaban M, Fawzy N. Image Classification Based On CNN: A Survey. Journal of Cybersecurity and Information Management (JCIM) 2021;6:18. https://doi.org/10.5281/zenodo.4897990.
[38] Purwono, Ma'arif A, Rahmaniar W, Fathurrahman HIK, Frisky AZK, Haq QMU. Understanding of Convolutional Neural Network (CNN): A Review. International Journal of Robotics and Control Systems 2022;2:739–48. https://doi.org/10.31763/ijrcs.v2i4.888.
[39] Sahu M, Dash R. A survey on deep learning: Convolution neural network (cnn). Smart Innovation, Systems and Technologies, vol. 153, Springer Science and Business Media Deutschland GmbH; 2021, p. 317–25. https://doi.org/10.1007/978-981-15-6202-0_32.
[40] Wang Z, Zhu S, Li J, Jiang W, Ramakrishnan KK, Yan M, Zhang X, Liu AX. DeepScaling: Autoscaling Microservices With Stable CPU Utilization for Large Scale Production Cloud Systems. IEEE/ACM Transactions on Networking 2024;32:3961–76. https://doi.org/10.1109/TNET.2024.3400953.
[41] Hu W, Hu G, Balakrishnan M, Yu X. Marlin: Efficient Coordination for Autoscaling Cloud DBMS. Proceedings of the ACM on Management of Data 2025;3:1–28. https://doi.org/10.1145/3749173.
[42] Roy N, Dubey A, Gokhale A. Efficient Autoscaling in the Cloud using Predictive Models for Workload Forecasting. n.d.
[43] Chen T, Bahsoon R, Yao X. A survey and taxonomy of self-aware and self-adaptive cloud autoscaling systems. ACM Comput Surv 2019;51. https://doi.org/10.1145/3190507.
[44] Semerikov S, Zubov D, Kupin A, Kosei M, Holiver V. Models and Technologies for Autoscaling Based on Machine Learning for Microservices Architecture. 2024.
[45] Nunes JPKS, Nejati S, Sabetzadeh M, Nakagawa EY. Self-adaptive, Requirements-driven Autoscaling of Microservices. Proceedings - 2024 IEEE/ACM 19th Symposium on Software Engineering for Adaptive and Self-Managing Systems, SEAMS 2024, Association for Computing Machinery, Inc; 2024, p. 168–74. https://doi.org/10.1145/3643915.3644094.
[46] Mampage A, Karunasekera S, Buyya R. A Deep Reinforcement Learning based Algorithm for Time and Cost Optimized Scaling of Serverless Applications 2023.
[47] Dashtbani M, Tahvildari L. Key Considerations for Auto-Scaling: Lessons from Benchmark Microservices 2025.
[48] Al-Dulaimy A, Taheri J, Kassler A, Hoseinyfarahabady MR, Deng S, Zomaya A. MultiScaler: A Multi-Loop Auto-Scaling Approach for Cloud-Based Applications. IEEE Transactions on Cloud Computing 2022;10:2769–86. https://doi.org/10.1109/TCC.2020.3031676.
[49] Ginarsa NAN, Santoso BJ. Intelligent Kubernetes Autoscaling Through Generative AI-Driven Workload Predictions. Proceedings - 2025 4th International Conference on Electronics Representation and Algorithm: Artificial Intelligence: Creating Tomorrow's World Today, ICERA 2025, Institute of Electrical and Electronics Engineers Inc.; 2025, p. 400–4. https://doi.org/10.1109/ICERA66156.2025.11087276.
[50] Valkenborg D, Geubbelmans M, Rousseau AJ, Burzykowski T. Supervised learning. American Journal of Orthodontics and Dentofacial Orthopedics 2023;164:146–9. https://doi.org/10.1016/j.ajodo.2023.04.010.
[51] Santos J, Reppas E, Wauters T, Volckaert B, De Turck F. Can Reinforcement Learning be Generalized for Efficient Auto-Scaling in Containerized Clouds? Proceedings of IEEE/IFIP Network Operations and Management Symposium 2025, NOMS 2025, Institute of Electrical and Electronics Engineers Inc.; 2025. https://doi.org/10.1109/NOMS57970.2025.11073717.
[52] Robino L, Garí Y, Pacini E, Mateos C, Yannibelli V, Monge DA. Reinforcement Learning-based Cloud Autoscaler Initialization via Evolutionary Algorithms 2025. https://doi.org/10.1145/3712255.

[53] Pandey V. Cost-Optimized Predictive Autoscaling of Cloud Resources Using Game-Theoretic Queuing Theory. Conference Proceedings - IEEE SOUTHEASTCON, Institute of Electrical and Electronics Engineers Inc.; 2025, p. 481–7. https://doi.org/10.1109/SoutheastCon56624.2025.10971700.
[54] Jafarnejad Ghomi E, Rahmani AM, Qader NN. Applying queue theory for modeling of cloud computing: A systematic review. Concurr Comput 2019;31. https://doi.org/10.1002/cpe.5186.
[55] Ding Y, Li C, Cai Z, Wang X, Yang B. A Dynamic Interval Auto-Scaling Optimization Method Based on Informer Time Series Prediction. IEEE Access 2025;13:14572–83. https://doi.org/10.1109/ACCESS.2024.3513564.
[56] Fog JW, Moller JJT, Jensen TM, Taibi D, Albano M. Comparing Neural and Statistical Time-Series Models for Proactive Auto-Scaling in Kubernetes. Proceedings - 19th IEEE International Conference on Service-Oriented System Engineering, SOSE 2025, Institute of Electrical and Electronics Engineers Inc.; 2025, p. 151–61. https://doi.org/10.1109/SOSE67019.2025.00022.
[57] Wang X, Jin Y, Schmitt S, Olhofer M. Recent Advances in Bayesian Optimization. ACM Comput Surv 2023;55. https://doi.org/10.1145/3582078.
[58] Zhang Y, Zhang F, Yang Z, Wang Z. What and How does In-Context Learning Learn? Bayesian Model Averaging, Parameterization, and Generalization 2023.
[59] Bazrafshan O, Ehteram M, Dashti Latif S, Feng Huang Y, Yenn Teo F, Najah Ahmed A, El-Shafie A. Predicting crop yields using a new robust Bayesian averaging model based on multiple hybrid ANFIS and MLP models: Predicting crop yields using a new robust Bayesian averaging model. Ain Shams Engineering Journal 2022;13. https://doi.org/10.1016/j.asej.2022.101724.
[60] Fragoso TM, Bertoli W, Louzada F. Bayesian Model Averaging: A Systematic Review and Conceptual Classification. International Statistical Review 2018;86:1–28. https://doi.org/10.1111/insr.12243.
[61] Khan ME. The Bayesian Learning Rule. vol. 24. 2023.
[62] Hinne M, Gronau QF, van den Bergh D, Wagenmakers EJ. A Conceptual Introduction to Bayesian Model Averaging. Adv Methods Pract Psychol Sci 2020;3:200–15. https://doi.org/10.1177/2515245919898657.
[63] Arbel J, Pitas K, Vladimirova M, Fortuin V. A Primer on Bayesian Neural Networks: Review and Debates 2023.
[64] Aslanpour MS, Toosi A, Cheema M, Chhetri M, Salehi M. Load balancing for heterogeneous serverless edge computing: A performance-driven and empirical approach. Future Generation Computer Systems 2024;154:266–80. https://doi.org/10.1016/j.future.2024.01.020.
[65] Chitsaz B, Khonsari A, Moradian M, Dadlani A, Talebi MS. Scaling Power Management in Cloud Data Centers: A Multi-Level Continuous-Time MDP Approach 2023. https://doi.org/10.48550/arXiv.2108.01292.
[66] Govindarajan K, Tienne AD. Resource Management in Serverless Computing: Review, Research Challenges, and Prospects. 12th International Conference on Advanced Computing (ICoAC), Chennai, India: IEEE; 2023. https://doi.org/10.1109/ICoAC59537.2023.10249574.
[67] Han R, Guo L, Ghanem MM, Guo Y. Lightweight Resource Scaling for Cloud Applications. 12th IEEE/ACM International Symposium on Cluster, Cloud and Grid Computing (CCGrid), 2012. https://doi.org/10.1109/CCGrid.2012.52.
[68] Koperek P, Funika W. Dynamic Business Metrics-Driven Resource Provisioning in Cloud Environments. Lecture Notes in Computer Science, vol. 7204, Springer; 2012, p. 171–80. https://doi.org/10.1007/978-3-642-31500-8_18.
[69] Anselmi J, Gaujal B, Rebuffi LS. Non-Stationary Gradient Descent for Optimal Auto-Scaling in Serverless Platforms. IEEE Transactions on Networking 2025;33:1574–87. https://doi.org/10.1109/TON.2025.3538982.
[70] Lorido-Botran T, Miguel-Alonso J, Lozano JA. Comparison of Auto-scaling Techniques for Cloud Environments 2013.
[71] Maurer M, Brandic I, Emeakaroha V, Dustdar S. Enacting SLAs in Clouds Using Rules. Euro-Par 2011: Parallel Processing, 2011, p. 147–52.
[72] Mampage A, Karunasekera S, Buyya R. Deadline-Aware Dynamic Resource Management in Serverless Computing Environments. 21st IEEE/ACM International Symposium on Cluster, Cloud and Internet Computing (CCGrid), IEEE; 2021, p. 483–92. https://doi.org/10.1109/CCGrid51090.2021.00058.
[73] Dutreilh X, Rivierre N, Moreau A, Malenfant J, Truck I. From Data Center Resource Allocation to Control Theory and Back. 2010 IEEE 3rd International Conference on Cloud Computing, 2010. https://doi.org/10.1109/CLOUD.2010.55.
[74] Lim HC, Babu S, Chase JS. Automated Control for Elastic Storage. 7th USENIX Conference on Networked Systems Design and Implementation (NSDI), 2010.

[75] Gambi A, Toffetti G. Modeling Cloud Performance with Kriging. 34th International Conference on Software Engineering (ICSE), 2012. https://doi.org/10.1109/ICSE.2012.6227075.
[76] Schuler L, Jamil S, Kühl N. AI-based Resource Allocation: Reinforcement Learning for Adaptive Auto-scaling in Serverless Environments 2020.
[77] Rao J, Bu X, Xu C-Z, Wang L, Yin G. VCONF: A Reinforcement Learning Approach to Virtual Machines Auto-Configuration. 6th International Conference on Autonomic Computing (ICAC), New York, NY, USA: ACM; 2009, p. 137–46. https://doi.org/10.1145/1555228.1555263.
[78] Zhang Z, Wang T, Li A, Zhang W. Adaptive Auto-Scaling of Delay-Sensitive Serverless Services with Reinforcement Learning. Proceedings - 2022 IEEE 46th Annual Computers, Software, and Applications Conference, COMPSAC 2022, Institute of Electrical and Electronics Engineers Inc.; 2022, p. 866–71. https://doi.org/10.1109/COMPSAC54236.2022.00137.
[79] Kumar J, Goomer R, Singh AK. Long Short Term Memory Recurrent Neural Network (LSTM-RNN) Based Workload Forecasting Model for Cloud Datacenters. Procedia Comput. Sci., vol. 125, 2018, p. 676–82. https://doi.org/10.1016/j.procs.2017.12.087.
[80] Anonymous. Automatic Cloud Resource Scaling Algorithm based on Long Short-Term Memory Recurrent Neural Network. International Journal of Advanced Computer Science and Applications 2016;7. https://doi.org/10.14569/ijacsa.2016.071236.
[81] Villela D, Pradhan P, Rubenstein D. Provisioning Servers in the Application Tier for E-commerce Systems. 12th International Workshop on Quality of Service (IWQoS), 2004. https://doi.org/10.1109/iwqos.2004.1309357.
[82] Salah K, Elbadawi K, Boutaba R. An Analytical Model for Estimating Cloud Resources of Elastic Services. Journal of Network and Systems Management 2016;24:285–308. https://doi.org/10.1007/s10922-015-9352-x.
[83] Nadjaran Toosi A, Son J, Chi Q, Buyya R. ElasticSFC: Auto-scaling techniques for elastic service function chaining in network functions virtualization-based clouds. Journal of Systems and Software 2019;152:108–19. https://doi.org/10.1016/j.jss.2019.02.052.
[84] Sfakianakis Y, Marazakis M, Kozanitis C, Bilas A. LatEst: Vertical Elasticity for Millisecond Serverless Execution. 22nd IEEE/ACM International Symposium on Cluster, Cloud and Internet Computing (CCGrid), 2022, p. 879–85. https://doi.org/10.1109/CCGrid54584.2022.00105.
[85] Khatua S, Ghosh A, Mukherjee N. Optimizing the Utilization of Virtual Resources in Cloud Environment. 2010 IEEE International Conference on Virtual Environments, Human-Computer Interfaces and Measurement Systems (VECIMS), 2010, p. 82–7. https://doi.org/10.1109/VECIMS.2010.5609349.
[86] Fang W, Lu ZH, Wu J, Cao ZY. RPPS: A Novel Resource Prediction and Provisioning Scheme in Cloud Data Center. 2012 IEEE 9th International Conference on Services Computing (SCC), 2012, p. 609–16. https://doi.org/10.1109/SCC.2012.47.
[87] Zhao L, Yang Y, Li Y, Zhou X, Li K. Understanding, predicting and scheduling serverless workloads under partial interference. Proceedings of the International Conference for High Performance Computing, Networking, Storage and Analysis, New York, NY, USA: Association for Computing Machinery; 2021. https://doi.org/10.1145/3458817.3476215.
[88] Almeida V, Arlitt M, Rolia J. Analyzing a web-based system's performance measures at multiple time scales. ACM SIGMETRICS Performance Evaluation Review 2002;30:3–9. https://doi.org/10.1145/588160.588162.
[89] Golshani E, Ashtiani M. Proactive auto-scaling for cloud environments using temporal convolutional neural networks. J Parallel Distrib Comput 2021;154:119–41. https://doi.org/10.1016/j.jpdc.2021.04.006.
[90] Taha MB, Sanjalawe Y, Al-Daraiseh A, Fraihat S, Al-E'mari SR. Proactive Auto-Scaling for Service Function Chains in Cloud Computing Based on Deep Learning. IEEE Access 2024;12:38575–93. https://doi.org/10.1109/ACCESS.2024.3375772.
[91] Javeed A, Borg A, Grahn H, Lundberg L, Patel D, Shirinbab S. Improving Cloud Efficiency: A Machine Learning-Based Stacking Model for CPU Utilization Prediction. Proceedings - 2025 8th International Conference on Data Science and Machine Learning Applications, CDMA 2025, Institute of Electrical and Electronics Engineers Inc.; 2025, p. 120–5. https://doi.org/10.1109/CDMA61895.2025.00026.
[92] Ouhame S, Hadi Y, Ullah A. An Efficient Forecasting Approach for Resource Utilization in Cloud Data Center Using CNN-LSTM Model. Neural Comput Appl 2021;33:10043–55. https://doi.org/10.1007/s00521-021-05770-9.

[93] Sabyasachi AS, Sahoo BM, Ranganath A. Deep CNN and LSTM Approaches for Efficient Workload Prediction in Cloud Environment. Procedia Comput Sci 2024;235:2651–61. https://doi.org/https://doi.org/10.1016/j.procs.2024.04.250.
[94] Vaghasia P, Goswami A, Patel D, Patel R, Patel R, Vaghasia R. Improving Predictive Accuracy with Cloud-Based Machine Learning Models for Big Data Analytics. 2025 International Conference on Computing Technologies, ICOCT 2025, Institute of Electrical and Electronics Engineers Inc.; 2025. https://doi.org/10.1109/ICOCT64433.2025.11118785.